\documentclass[preprint,12pt,authoryear]{elsarticle}
\usepackage{gensymb}
\usepackage{amssymb}
\usepackage{multirow}
\usepackage{caption}
\usepackage{booktabs}
\usepackage{epsfig}
\usepackage{longtable}
\usepackage{rotating}
\usepackage{amsmath}
\usepackage{array}
\usepackage[bottom]{footmisc}
\usepackage[hyphens]{url} 
\usepackage[colorlinks,urlcolor=blue]{hyperref}
\usepackage{hyperref}
\usepackage{color, colortbl}
\usepackage{subfigure}
\makeatletter 
\def\fullwidthdisplay{\displayindent\z@ \displaywidth\columnwidth}
\edef\@tempa{\noexpand\fullwidthdisplay\the\everydisplay}
\everydisplay\expandafter{\@tempa}
\makeatother

\definecolor{Gray}{gray}{0.9}

\newcommand{\edit}[1]{\color{red}{#1}\color{black}{}}
\frontmatter          
\journal{NeuroImage}

\begin{document}

\begin{frontmatter}

\title{Fully Convolutional Network Ensembles for White Matter Hyperintensities Segmentation in MR Images}


\author{Hongwei~Li$^{1,2,3}$, Gongfa~Jiang$^{1}$, Jianguo~Zhang$^{2}$, Ruixuan~Wang$^{1}$, Zhaolei~Wang$^{1}$,
Wei-Shi~Zheng$^{1}$ and Bjoern~Menze$^{3}$}

\address{1. School of Data and Computer Science, Sun Yat-sen University, China.\\
         2. Computing, School of Science and Engineering, University of Dundee, UK.\\
         3. Department of Computer Science, Technical University of Munich, Germany.}

\begin{abstract}
White matter hyperintensities (WMH) are commonly found in the brains of healthy elderly individuals and have been associated with various neurological and geriatric disorders. 
In this paper, we present a study using deep fully convolutional network and ensemble models to automatically detect such WMH using fluid attenuation inversion recovery (FLAIR) and T1 magnetic resonance (MR) scans.
The algorithm was evaluated and ranked 1$^{st}$ in the WMH Segmentation Challenge at MICCAI 2017.
In the evaluation stage, the implementation of the algorithm was submitted to the challenge organizers, who then independently tested it on a hidden set of 110 cases from 5 scanners. 
Averaged dice score, precision and robust Hausdorff distance obtained on held-out test datasets were 80\%, 84\% and 6.30mm respectively. These were the highest achieved in the challenge, suggesting the proposed method is the state-of-the-art.
Detailed descriptions and quantitative analysis on key components of the system were provided.
Furthermore, a study of cross-scanner evaluation is presented to discuss how the combination of modalities
affect the generalization capability of the system. The adaptability of the system to different scanners and protocols is also investigated. A quantitative study is further presented to show the effect of ensemble size and the effectiveness of the ensemble model. Additionally, software and models of our method are made publicly available.
The effectiveness and generalization capability of the proposed system show its potential for real-world clinical practice.
\end{abstract}


\end{frontmatter}


\section{Introduction}
Small vessel diseases are mainly systemic disorders that affect various tissues and organs of human body.
These diseases are thought to be the most frequent pathological neurological process and have a crucial role in at least three fields: stroke, dementia and aging \citep{pantoni2010cerebral}.

\begin{figure}[t]
	\begin{center}
		\vspace{0.2cm}
		\includegraphics[width=0.75\linewidth,height=0.35\linewidth]{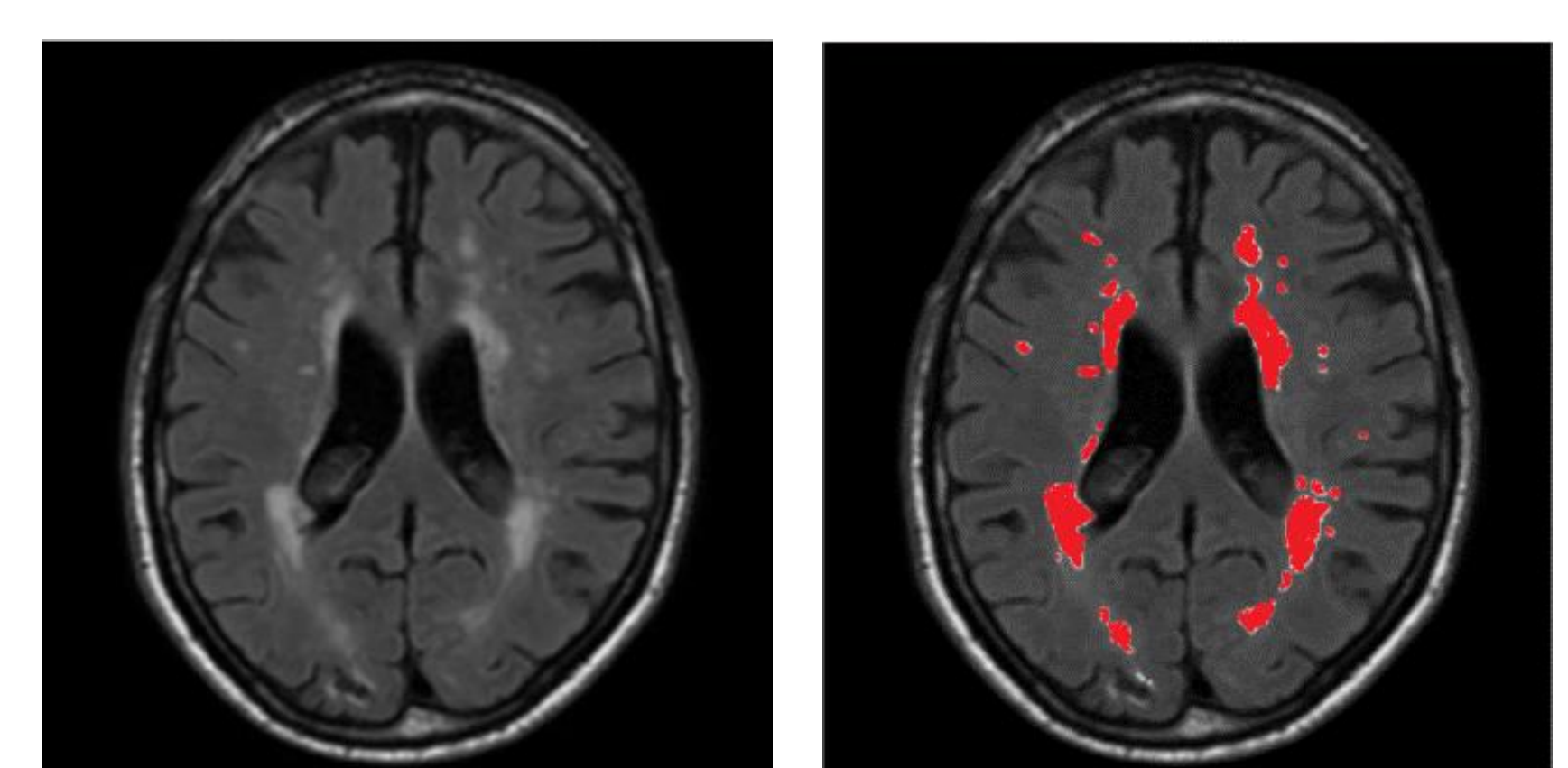}
	\end{center}
	\caption{A sample of MR slice from FLAIR modality (left), and its corresponding manual annotation of WMH by a neuroradiologist (right).}
	\label{fig:figure1} 
\end{figure}

White matter lesions characterized by bilateral, mostly symmetrical hyperintensities\edit{,} are commonly seen on FLAIR MRI of clinically healthy elderly people; furthermore, they have been repeatedly associated with various neurological and geriatric disorders such as mood problems and cognitive decline \citep{kim2008classification, debette2010clinical}.
Manual delineation of WMH area, as shown in Figure \ref{fig:figure1}, is a reliable way to assess white matter abnormalities but this process is laborious and time-consuming for neuroradiologists and shows high intra-rater and inter-rater variability \citep{grimaud1996quantification}.

Computer vision and machine learning techniques have increasingly shown a promising road for automatic diagnosis of diseases through medical imaging. By analyzing imaging data in a statistical manner, many image processing algorithms dealing with brain lesions generalize well within closely related applications, for example, in the segmentation of WMH, multiple sclerosis (MS), tumors, stroke, and even traumatic brain injury.
Although various computer-aided diagnosis systems have been proposed for these different brain lesion segmentation tasks, the reported results are largely incomparable due to different datasets and evaluation protocols.

\cite{van2001automated}~presented an early attempt at developing an unsupervised-learning-based segmentation system to detect multiple sclerosis lesions from large datasets of T1-weighted (T1), proton density-weighted (PD) and T2-weighted (T2) scans. The method simultaneously estimates the parameters of a stochastic model for normal brain MR images and detects MS lesions as outliers of the model. \cite{anbeek2004probabilistic} developed a supervised-learning-based automated system using T1, inversion recovery, PD, T2 and fluid attenuation inversion recovery (FLAIR) scans. Intensity and 3D spatial features were extracted from the voxels and are used to train a k-nearest neighbors classifier.
\cite{dyrby2008segmentation} used artificial neural networks based on intensity and spatial information, in which six optimized networks were produced to investigate the impact of different input modalities on WMH segmentation. 
\cite{beare2009development} developed a method that searched for WMHs per-region instead of per-voxel. The region-based features are combined with an adaptive boosting statistical classifier.
\cite{geremia2010spatial, geremia2011spatial} were the first to address the MS lesion segmentation in a straightforward learning approach using context-rich, symmetry and local spacial features and random forest.
\cite{simoes2013automatic} built the intensity histogram of FLAIR by a Gaussian mixture model. Then the probability of a voxel depends on not only the voxel's intensity but also on its neighbors' current class probabilities.
\cite{schmidt2013lst} contributed an open source tool for the segmentation of hyperintensities that integrates with the popular SPM package.
\cite{yoo2014application} developed an intensity-based, monospectral segmentation method in which the optimal intensity threshold on FLAIR images varied with WMH volume.
Very recently, \cite{ghafoorian2017location} integrated the anatomical location information into the convolutional neural networks (CNN), in which several deep CNN architectures that consider multi-scale patches or take explicit location features were proposed.
\cite{moeskops2017evaluation} proposed a patch-based deep CNN to segment brain tissues and WMH in MR images.

In computing research, benchmarking on specific problems is an effective way to fairly compare state-of-the-art methods. There have been several related benchmarks on automated segmentation of different brain tissues in MR images in the field of medical image analysis. The Multiple Sclerosis Lesion Segmentation Challenge 2008 organized by \cite{styner20083d} is one of the early contests for comparing the methods for automatic extraction of MS lesions from T1, T2 and FLAIR MRI data.
The Ischemic Stroke Lesion Segmentation Challenge (ISLES) from 2015 to 2017 organized by \cite{maier2017isles} provides a platform for fair comparison of stroke lesion segmentation algorithms.
The Multi-modal Brain Tumor Segmentation Challenge (BRATS) organized by \cite{menze2015multimodal} draws much attention since 2012 which focuses on segmentation of low- and high-grade gliomas, more recently, prediction of patient overall survival. Different from the above tasks, WMH tend to have consistent patterns such as significant symmetry, but they are more scattered, often with some regions of very small size and irregular shapes.
Furthermore, compared with other brain tissue segmentations, WMH segmentations are more likely to be susceptible to the presence of motion artefacts and other brain abnormalities, such as brain infarcts \citep{gouw2010heterogeneity}.

The \emph{WMH Segmentation Challenge 2017} \footnote{\url{http://wmh.isi.uu.nl/}} was held to compare state-of-the-art algorithms in conjunction with the 20$^{th}$ International Conference on Medical Image Computing and Computer Assisted Intervention (MICCAI 2017).
This paper describes our winning entry to this challenge in detail, which was evaluated by the organizers on clinical datasets. The algorithm was containerized and applied to the test datasets by the challenge organizers, while the test sets remained unseen to us and other contestants. The test set includes 110 secret cases from five different MR scanners world-widely from three hospitals in the Netherlands and Singapore. Our approach to detecting WMH in MR images is based on an ensemble of convolution-deconvolution architecture \citep{long2015fully} with long-range connections \citep{ronneberger2015u} which simultaneously classifies each pixel and locates objects of an input image.
In our system, we implement a network architecture with 19 layers that are optimized for classifying and localizing the WMH. Ensemble models trained with random parameter initializations and shuffled data are employed for voting the pixel labels in the final evaluation.

This paper is organized as follows. Section \ref{materials} describes the datasets, rating criteria, five evaluation metrics on segmentation performance and rank method of the challenge. Section \ref{methods} presents in detail each component of our method and how some key parameters are optimized. Section \ref{experiments} evaluates the proposed system on the public training dataset (60 cases) and reports results for the hidden held-out dataset (110 cases). Section \ref{discussion} discusses different aspects of our winning method. This includes the motivation to use 2D model instead of 3D one, a novel \textit{cross-scanner} study on how the combination of modalities and data augmentation strengthen the generalization capability to unseen scanners. Furthermore, evaluation on the adaptability to various scanners as well as quantitative analysis on the optimal number of ensemble models are performed.

\label{}

\section{Materials} \label{materials}
This section mainly describes the WMH Segmentation Challenge, datasets, evaluation metrics and rank method which are referred to in the rest of the article.
\subsection{MICCAI WMH Segmentation Challenge Overview}
The challenge organized as a joint effort of the \emph{UMC Utrecht, VU Amsterdam} and \emph{NUHS Singapore}, aims at, for the first time, benchmarking methods for automatic WMH segmentation of presumed vascular origin. Sixty cases from three centers were released as a public training set for participants to build and evaluate their algorithms. One hundred and ten hidden cases from five scanners are used by the organizers to test the algorithms. Notably, all algorithms are containerized by \emph{Docker} \citep{merkel2014docker} to guarantee that the test data remains secret and cannot be included in any way in the training procedure of the techniques.
Twenty international teams participated, and further information including training data and the results on test set are made public via the following url: \url{http://wmh.isi.uu.nl/results/}.

\subsection{Datasets} \label{datasets}
In all reported experiments, we relied on the publicly available dataset from the MICCAI WMH Challenge. Properties of the data are summarised in Table \ref{table:Table2}.
A notable feature is that the images were acquired from five different scanners from three hospitals in the Netherlands and Singapore. As shown in Table \ref{table:Table2}, there exists large difference in acquisition settings; in particular voxel sizes of the captured images differ significantly among the five scanners.
For each subject, a $3D$ T1-weighted image, and a $2D$ multi-slice FLAIR image were provided.
Since the manual reference standard is defined on the FLAIR image, a $2D$ multi-slice version of the T1 image was generated by re-sampling the $3D$ T1-weighted image to match with the FLAIR one.
Finally, the pre-processed images were corrected for bias field inhomogeneities using \emph{SPM12}\footnote{\url{http://www.fil.ion.ucl.ac.uk/spm/software/spm12/}}. The 3D FLAIR image was resampled to a slice-thickness of 3.00 mm and there is no gap between slices.

The dataset consists of in total 170 subjects with FLAIR and T1 MR images from five different scanners along with their binary masks.
The images from 60 subjects were made available during the training stage. The images from the remaining 110 subjects were used as the hidden test set to evaluate performance of methods submitted to the challenge.
Notably, the test set also includes images of 20 subjects captured by other two \textit{unseen} scanners, which were not used to capture images for training.
This dataset setting encourages the participants to submit algorithms that could be robust to unseen scanners.

\begin{table*}[t]
	\vspace{0.2cm}
	\scriptsize
	\newcommand{\tabincell}[2]{\begin{tabular}{@{}#1@{}}#2\end{tabular}}
	\renewcommand\arraystretch{1}
	\centering
	\caption{Characteristics of \textit{MICCAI WMH Challenge} dataset. The training set consists 60 subjects' data from 3 scanners and the test set includes 110 cases from 5 scanners (two of them are not represented in the training set)}\label{table:Table2}.
	\begin{tabular}{clccccc}
		\toprule
		\textbf{Datasets}&\textbf{Scanners Name}&\textbf{Voxel Size $(m^3)$} &\textbf{Size of FLAIR Scans}&\textbf{Train}&\textbf{Test}\\
		\midrule
		\emph{Utrecht}&3T Philips Achieva & 0.96$\times$0.95$\times$3.00 & 240$\times$240$\times$48 &20 &\textbf{30}\\
		\emph{Singapore}&3T Siemens TrioTim & 1.00$\times$1.00$\times$3.00 & 252$\times$232$\times$48 &20 &\textbf{30}\\
		\emph{GE3T}&3T GE Signa HDxt& 0.98$\times$0.98$\times$1.20&132$\times$256$\times$83 &20 &\textbf{30}\\ \midrule
		{GE1.5T}&3T Philips Ingenuity & 1.04$\times$1.04$\times$0.56 & secret &- &\textbf{10}\\
	   {PETMR}&1.5T GE Signa HDxt & 1.21$\times$1.21$\times$1.30& secret &- &\textbf{10}\\
		\bottomrule
	\end{tabular}
\label{tab:datasets}
\end{table*}

\subsection{Evaluation Metrics and Rank Method}\label{evaluationAndRank}
Five different metrics are used by the challenge organizers to compare and rank the methods by different teams; those metrics evaluate the segmentation performance in different aspects.

Given a ground-truth segmentation map $G$ and a segmentation map $P$ generated by an algorithm, the five evaluation metrics are defined as follows.
\subsubsection{Dice similarity coefficient (DSC)}
	\begin{equation}
	\emph{DSC} = \frac{2(G\cap{P})}{|G|+|P|}
	\end{equation}
	This measures the overlap in percentage between $G$ and $P$.

\subsubsection{Hausdorff distance $($95$^{th}$ percentile$)$}
Hausdorff distance is defined as:
	\begin{equation}
	\emph{$H(G,P)$} = max\{\sup\limits_{x\in G} \inf\limits_{y\in P} d(x,y), \sup\limits_{y\in P} \inf\limits_{x\in G} d(x,y)\}
	\end{equation}
	where \emph{d(x, y)} denotes the distance of \emph{x} and \emph{y}, \emph{sup} denotes the supremum and \emph{inf} for the infimum.
	This measures how far two subsets of a metric space are from each other. As used in this challenge, it is modified to obtain a robustified version by using the 95$^{th}$ percentile instead of the maximum (100$^{th}$ percentile) distance.

\subsubsection{Average volume difference (in percentage)}
Let $V_{G}$ and $V_{P}$ be the volume of lesion regions in $G$ and $P$ respectively.
	Then the Average Volume Difference (AVD) in percentage is defined as:
	\begin{equation}
	\emph{AVD} = \frac{|V_{G}-V_{P}|}{V_{G}}
	\end{equation}
\subsubsection{Sensitivity for individual lesions (recall)}
	Let $N_{G}$ be the number of individual lesions delineated in $G$, and $N_{P}$ be the number of correctly detected lesions after comparing $P$ to $G$. Each individual lesion is defined as a 3D connected component.
	Then the recall for individual lesions is defined as:
	\begin{equation}
	\emph{Recall} = \frac{N_P}{N_G}
	\end{equation}
	
\subsubsection{F1-score for individual lesions}
	Let $N_{P}$ be the number of correctly detected lesions after comparing $P$ to $G$. $N_{F}$ be the number of wrongly detected lesions in $P$. Each individual lesion is defined as a 3D connected component.
	Then the F1-score for individual lesions is defined as:
	\begin{equation}
	\emph{F1} = \frac{N_P}{N_{P}+N_{F}}
	\end{equation}
	
The full source code for computing the evaluation metrics can be found on: \url{https://github.com/hjkuijf/wmhchallenge/blob/master/evaluation.py}.

For each team, the values of those five metrics were computed by the organizers independently.
For each evaluation metric, the performances of all of the teams were sorted from best to worst.  Then a calibrated score for each team was computed by normalising its performance w.r.t the range of all the actual performances for that metric. Thus the best team was assigned a rank score of one, while the worst team got a rank score of zero. Other teams received a score of between (0,1).
Finally, for each team, the rank scores of the five metric were averaged into the final score, being the overall performance of that team. For consistency, when presenting the results of the challenge, we follow exactly the same ranking criteria.
\section{Methods} \label{methods}

\begin{figure}[t]
	\begin{center}
		\includegraphics[width=1.02\linewidth,height=0.53\linewidth]{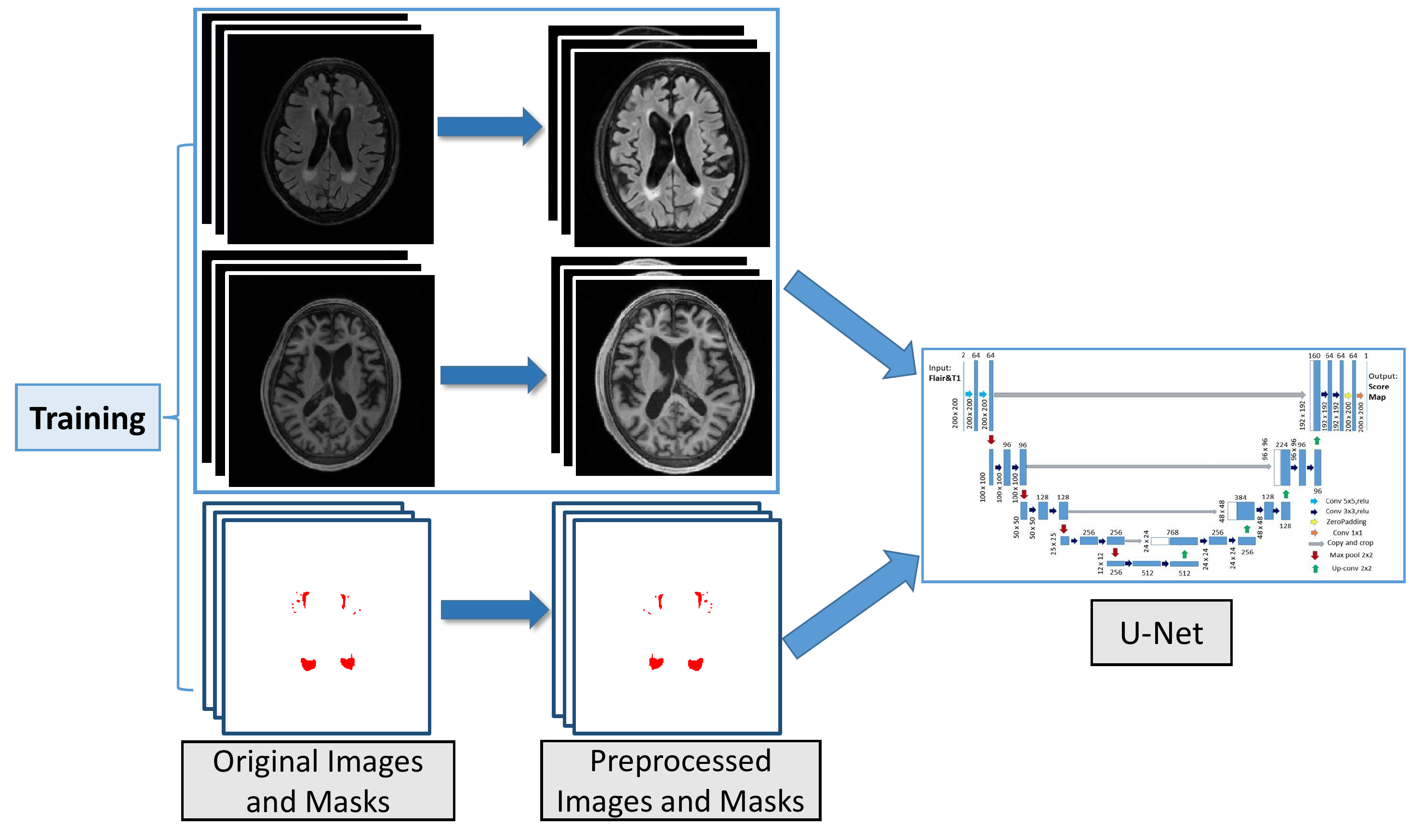}
	\end{center}
	\caption{Overall framework of the training stage.}
	\label{fig:segmentation} \vspace{-0.5cm}
\end{figure}

\subsection{Further Preprocessing} \label{preprocessing}
A further preprocessing on top of the basic preprocessing steps pursued by the organizers (Section \ref{datasets}) plays an important role in our overall framework. We aim at employing a simple and effective preprocessing step on both training and held-out testing set.
It is motivated by three objectives: 1) to guarantee a uniform size of all data for deep convolutional networks in the training and test stage, 2) to normalize voxel intensity to reduce variation across subjects.  and 3) to equip the CAD system with desired invariance and robustness. We enforce these desired data properties by implementing further steps in the training of our algorithm: 1) cropping or padding each axial slice to a uniform size, 2) Gaussian normalization on the brain voxel intensity, and 3) data augmentation on the processed images.
Most of these steps are performed for both FLAIR and T1 modalities and for both the training and test stages.
Data augmentation was performed only during the training stage.

Firstly, all the axial slices were automatically cropped or padded to $200\times200$, in order to guarantee a uniform size for input to the deep-learning model. 
Secondly, Gaussian normalization was employed to normalize the intensity distributions for each 3D scan.
This includes three steps. Firstly, a threshold was empirically set to obtain an initial binary brain mask.
Secondly, for each axial slice of the obtained binary masks, the largest connected component was selected. Thirdly, the holes inside the connected component was filled using morphology operations. Thus a final brain mask was obtained for each slice. For each 3D scan, Gaussian normalization was then employed to rescale the voxel intensities \textit{within} each individual's brain mask.

The thresholds for creating the brain masks were empirically set to 70 for FLAIR and 30 for T1 respectively. It was noted that several methods submitted for the contest extracted the brain using common tools such as BET \citep{smith2002fast}, where the skull was also removed. However, we found the removal of skull has little effect on the performance of the proposed system.

\subsubsection{Data augmentation}
Data augmentation is an effective way to equip the deep networks with desired invariance and robustness properties when training data are limited. In case of MR images among different subjects and scanners, due to variations of head orientations, voxel sizes and WMH distribution, we primarily need rotation and scale invariance as well as robustness to shear transformation.
For each axial slice, three transformations including rotation, shear mapping and scaling were applied, each within a parameter range. The parameter range represents the variation in different aspects between subjects in clinical practice; for example, rotation of brain is in the range of [-15\degree, 15\degree]. Table \ref{table:Table1} lists the parameter range for each of the three transformations. It should be noted that the scaling used in the training of the algorithm was in the range of (0.9, 1.1), representing the range of voxel size ratios in the training data sets (Table \ref{tab:datasets}), while some test sets had noticeable larger ratios (a factor of 1.21 between the PETMR and the Singapore data set). This indicates the robustness of our approach, but also leaves potential room for improvement in future studies exploring the optimal scaling of the data during training.

\begin{figure*}[t]
	\begin{center}
		\includegraphics[width=\linewidth,height=0.3\linewidth]{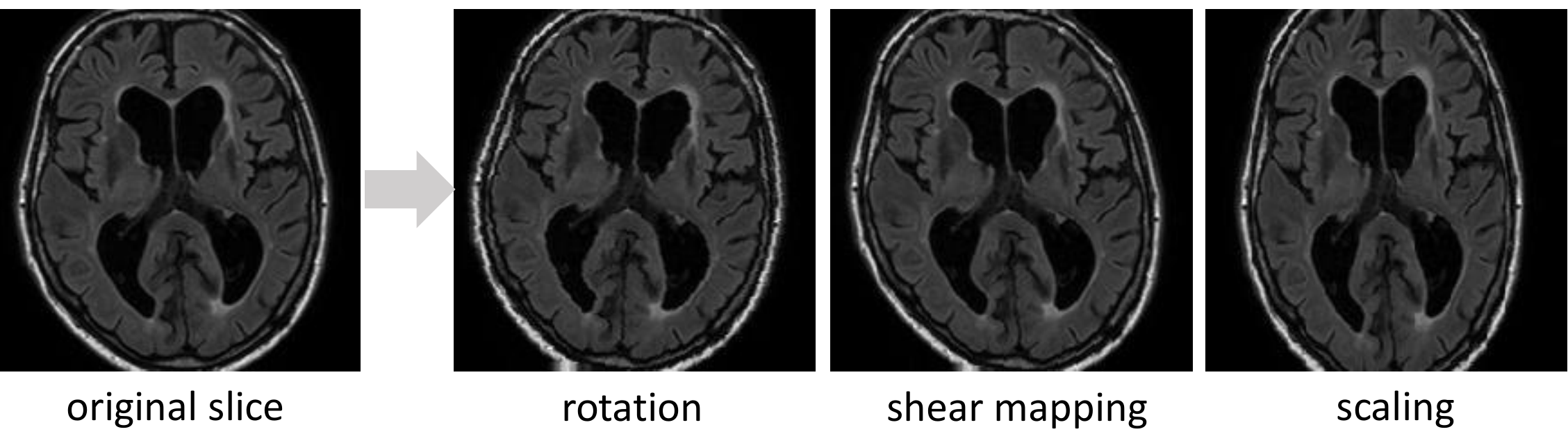}
	\end{center}
	\caption{An example of data augmentation result. From left to right: the original axial slice, slice after rotation, slice after shear mapping and slice after scaling.}
	\label{fig:dataAugmentation} 
\end{figure*}

Figure \ref{fig:dataAugmentation} shows an example of the resulting slices after applying the transformations. After data augmentation, we obtain a dataset ten times larger than the original one.
\begin{table}[h]
	\vspace{0.01cm}
	\scriptsize
	\newcommand{\tabincell}[2]{\begin{tabular}{@{}#1@{}}#2\end{tabular}}
	\renewcommand\arraystretch{1}
	\centering
\caption{Parameters range used for data augmentation. The value range in column \emph{Shearing} indicates the shear angle. The value range in column \emph{scaling} indicates the scale factor.}\label{table:Table1}
	\begin{tabular}{lccc}
		\toprule
		
		\textbf{Methods}&~~~\textbf{Rotation}&~~~~\textbf{Shearing}&~~~\textbf{Scaling $($x \& y$)$}\\
		\midrule
		\textbf{Parameters}&~~~[-15\degree, 15\degree]&~~~~[-18\degree, 18\degree] &~~~[0.9, 1.1]\\
		\bottomrule
	\end{tabular}
	\vspace{0.1cm}
	
\end{table}

\subsection{Fully Convolutional Network}

\subsubsection{2-D Convolutional Network Architecture}
\begin{figure*}[t]
	\begin{center}
		\includegraphics[width=\linewidth,height=0.40\linewidth]{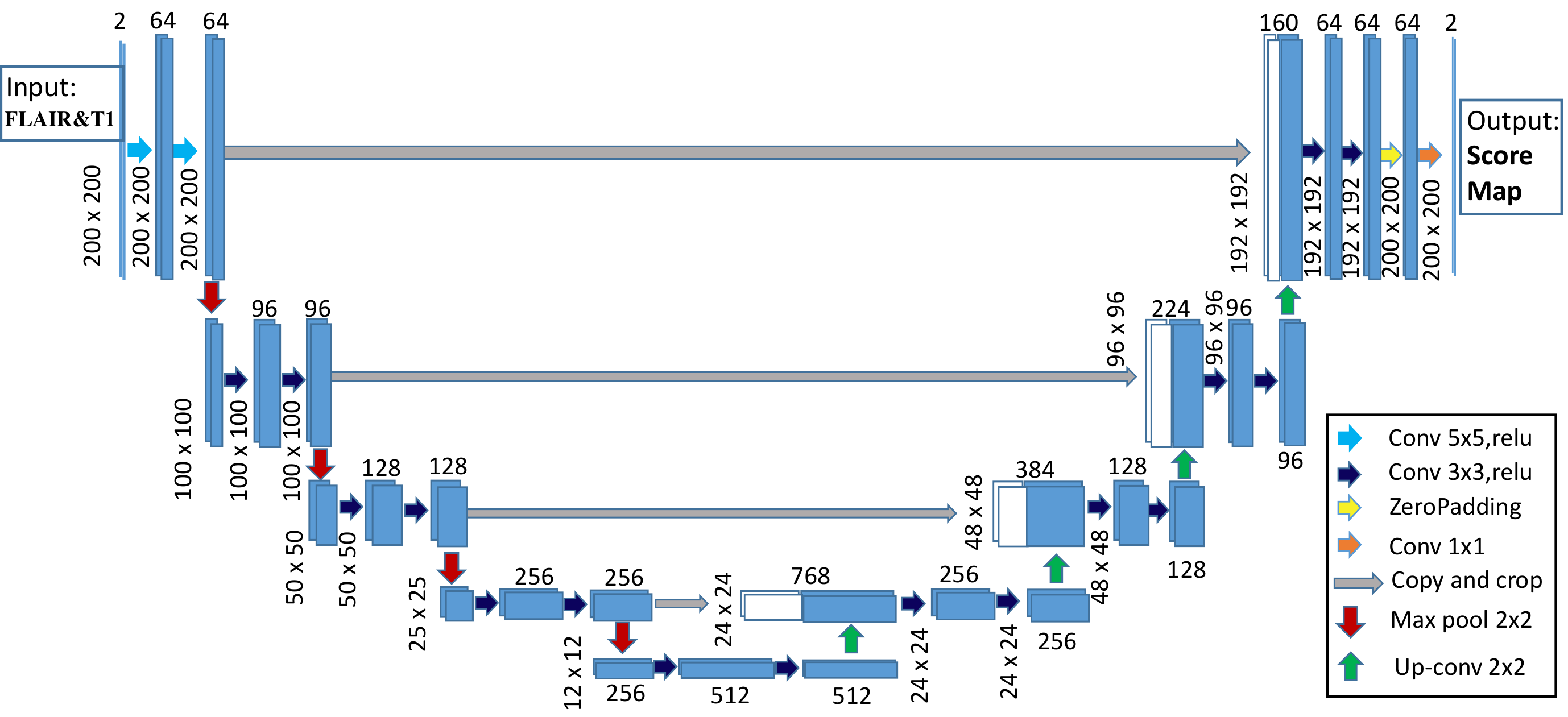}
	\end{center}
	\caption{2D Convolutional Network Architecture. It consists of a shrinking part (left side) and an expansive part (right side) to detect and locate\emph{WMH} respectively. The input includes FLAIR and T1 channel.}
	\label{fig:architectureUnet} 
\end{figure*}

Convolutional neural network has proven to be an effective computational model for automatically extracting image features. Recently the fully convolutional networks (FCN) \citep{long2015fully} and their its extensions \citep{milletari2016v} have been used for medical images segmentation.
We build a variant of FCN architecture based on U-Net \citep{ronneberger2015u}, which takes as input the axial slices of two modalities from the brain MR scans during both training and testing.  Our network is shown in Figure \ref{fig:architectureUnet}. For each patient, the FLAIR and T1 modalities are fed into the U-Net jointly as a two-channel input.
It consists of a down-convolutional part that shrinks the spatial dimensions (left side), and up-convolutional part that expands the score maps (right side).
The skip connections between down-convolutional and up-convolutional were employed.

In this model, two convolutional layers are repeatedly employed, each followed by a rectified linear unit (ReLU) and a 2$\times$2 max pooling operation with stride 2 for downsampling. At the final layer a 1$\times$1 convolution is used to map each 64-component feature vector to two classes.
In total the network contains 19 convolutional layers.
Convolutional layers with $3\times3$ kernel size are heavily used in our model.
Different from the basic architecture of the recent work \citep{ronneberger2015u},
for the first two convolutional layers, kernel size $3\times3$ is replaced with size $5\times5$ in order to handle different transformations. This is motivated by a recent study \citep{peng2017large} suggesting that large kernel size should be adopted in the network architecture. This step could enable dense connections between feature maps and per-pixel classifiers, enhancing the capability of a network to handle different transformations.

\subsubsection{Dice Loss}
In the task of WMH segmentation, the numbers of positives and negatives are highly unbalanced.
One of the solutions to tackle this issue is to use Dice loss \citep{milletari2016v} as the loss function for training the model.
The formulation is as follows.

Let G = \{g$_{1}$, ..., g$_{N}$\} be the ground-truth segmentation probabilistic maps (gold standard) over $N$ slices, and P = \{p$_{1}$, ..., p$_{N}$\} be the predicted probabilistic maps over $N$ slices.
The Dice loss function can be expressed as:

\begin{equation}
\emph{DL} =  - \frac{2\sum_{n = 1}^N | p_{n} \circ g_{n}| + s}{\sum_{n = 1}^N (|p_{n}| + |g_{n}|) + s}
\end{equation}

where $\circ$ represents the entrywise product of two matrices, and $|\cdot|$ represents the sum of the entries of matrix. The \emph{s} term is used here to ensure the loss function stability by avoiding the division by $0$, i.e., in a case where the entries of $G$ and $P$ are all zeros. \emph{s} was set to 1 in our experiments.

%

\subsection{Ensemble FCNs}
Ensemble techniques are helpful to reduce over-fitting problems of a complex model on the training data \citep{opitz1999popular}.
It combines multiple learning models to obtain better predictive performance than any of the constituent learning algorithms alone.
There exists various work using ensembles of deep learning models in computer vision and medical image analysis. \cite{krizhevsky2012imagenet} and \cite{simonyan2014very} achieved top performance in the ImageNet Large Scale Visual Recognition Challenge (ILSVRC) 2012 and 2014 by averaging multiple deep CNNs with same architectures.
\cite{he2016deep} won the first place with an ensemble of six Residual Networks with different depths in ILSVRC 2015. \cite{kamnitsas2017ensembles} won the brain tumor segmentation challenge (BraTs) 2016 by aggregating different segmentation networks. 
In this work, we propose to address the automated WMH segmentation problem by an ensemble approach to combine several models with same architecture in a carefully designed pipeline. We further show the effectiveness of the ensemble model via a quantitative analysis in Sections \ref{ensembles} and \ref{sec:analysis}.


The intention to use ensemble models includes two aspects: 1) different models could learn different attributes of the training data during the batch learning processing, thus the ensemble of them could boost the segmentation results; 2) bias-variance trade-off. Assume that network model error is due to bias and variance. If the variance of model decrease, then the overall error would likely decrease. Here we aimed to lower the variance by averaging the model outputs. A FCN with millions of parameters, over-trained on different bootstrapped/subsampled training sets would qualify for unbiased and highly variant models. We further discussed in Section \ref{ensembles} that ensemble model served as the typical bias-variance trade-off.

\begin{figure*}[t]
	\begin{center}
		\includegraphics[width=\linewidth,height=0.40\linewidth]{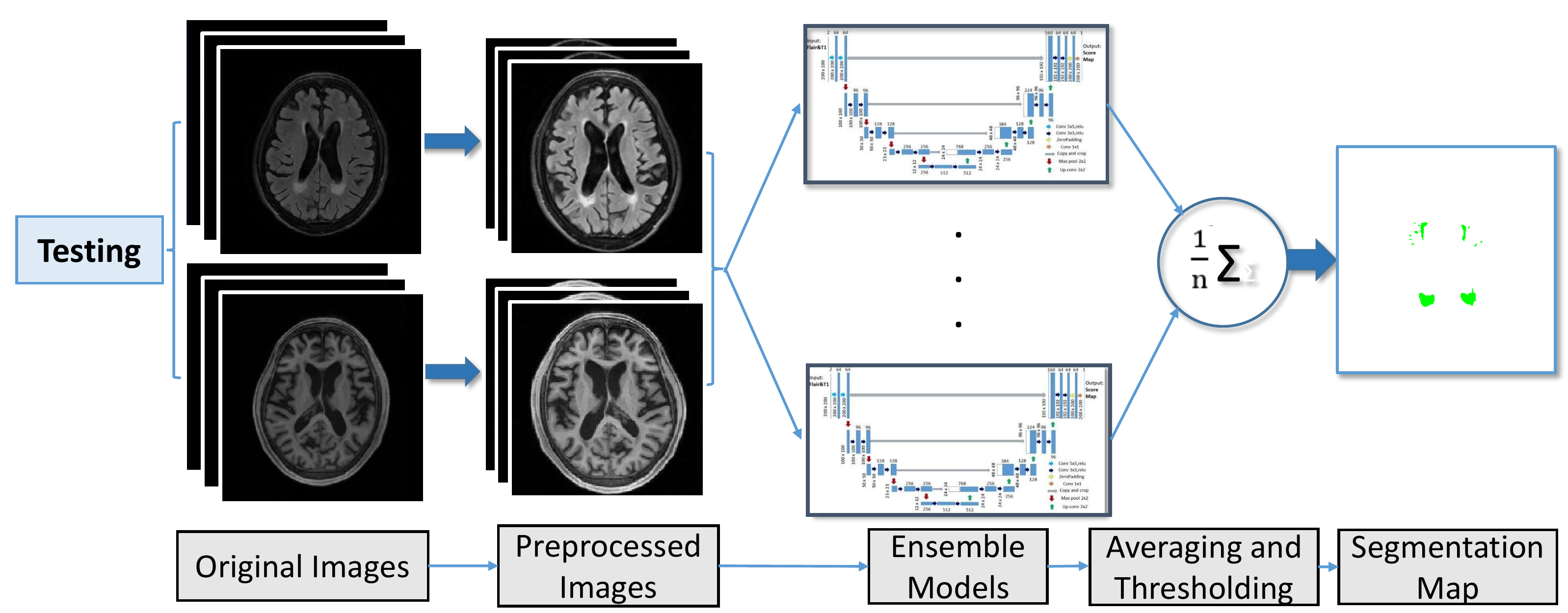}
	\end{center}
	\caption{Overall framework for the testing stage.}
	\label{fig:test_system} 
\end{figure*}

As shown in Figure \ref{fig:test_system}, $n$ U-Net models with same architecture are trained with random parameter initialization and shuffled data in the batch learning. For each of the $n$ U-Net models, when given a test image, a probability segmentation map will be generated by that model.
Then the resulting $n$ maps will be averaged. Finally an empirically-picked threshold will be used to transform the scores map into a binary segmentation map.

\subsection{Post-processing}
The post-processing includes two aspects: 1) cropping or padding the segmentation maps with respect to the original size, i.e., an inverse operation to the step described in Section \ref{preprocessing}; 2) removing some anatomically unreasonable artefact in the axial slices.
For the purpose of removing unreasonable detections (e.g., WMH will not appear in the first few axial slices containing neck and last few axial slices containing skull), we employed a simple strategy: if there exists detected WMH in the first $m$ slices and last $n$ ones of a brain along the $z$-direction, then the WMH regions were considered as false positive and would be removed. Empirically, $m$ and $n$ were set to 10\% of the number of slices for each scan.
The codes and models of the proposed system is made publicly available in \emph{GitHub}\footnote{\url{https://github.com/hongweilibran/wmh_ibbmTum}}.

\section{Results} \label{experiments}
In this section we report the segmentation performances on both the public training dataset and the held-out test set and compare to other teams' methods presented during the challenge. Detailed segmentation results of the 20 teams on the 110 secret cases are available in the following url:
\url{http://wmh.isi.uu.nl/results/}.

For reported results, the binary segmentation maps were evaluated using the five metrics described in Section \ref{materials}: dice similarity coefficient, Hausdorff distance (95p), averaged volume difference, lesions recall and lesions F1-score. The U-Net hyper parameters were set as follows: batch size for computing the training loss was set to 30; learning rate was set to 0.0002; the number of epochs was set to 50. The number of models in the ensemble was set to 3. Section \ref{parameters} further evaluates and analyses the effects of some key parameters on segmentation performance.

\subsection{Results on held-out test dataset}

The proposed system was announced to be the winning method of the challenge after being independently tested on 110 hidden cases from 5 scanners by the organizers. The overall ranking was based on the average of the rank scores computed for each metric. For the testing stage, deep fully convolutional networks were learned on the whole public training dataset consisting of 60 cases.
Table \ref{fig:testSetResultFiveSets} shows the segmentation performance of our submitted system on the held-out test set with its 5 subsets, each containing cases from the different scanners and sites. Table \ref{fig:testSetResultComparsion} compares our method to other top performing teams. Notably, the top-5 methods all used deep learning techniques, briefly described in Table \ref{table:Table8}.
The proposed FCN ensemble achieved, on average, the highest dice similarity coefficient, smallest Hausdorff distance and best lesion recall.
For the 20 cases from unseen scanners \emph{AMS GE1.5T} and \emph{AMS PETMR}, our method achieved the highest lesion recalls of 90\% and 84\% respectively.
We will discuss in Section \ref{discussion} how each key component of our method, especially the model ensemble, contributes to the improvement on the generalization capability. 


\begin{table}
    \caption{Results of our method on the heldout sets from the five different scanners. $\downarrow$ indicates that smaller value represents better performance. The last row shows the rank scores of our method w.r.t the 20 teams for each of the five metrics, with  0=\emph{best}, and 1=\emph{worst}.}
    \begin{tabular}{ | c | c | c | c | c | c |}
    \hline
     \textbf{Scanners}&\textbf{~DSC~} & \textbf{~H95~$\downarrow$} & \textbf{~AVD~$\downarrow$} & \textbf{Recall} & ~~\textbf{F1}~\\
    \hline
     \emph{Utrecht (n = 30)}   & 0.80 & 7.22  & 18.35  & 0.81  & 0.72\\
    \hline
     \emph{Singapore (n = 30)} & 0.83 & 4.50  & 19.95  &  0.85  & 0.78\\
    \hline
     \emph{GE3T (n = 30)}      & 0.79 &  4.04 & 24.46 & 0.83  & 0.79\\
    \hline
     \emph{AMS GE1.5T (n = 10)}& 0.77 &  10.24 & 36.86 & 0.90 & 0.80\\
    \hline
     \emph{AMS PETMR (n = 10)} & 0.72 &  11.84 & 15.54 & 0.84  & 0.65\\
    \hline
     weighted average          & \textbf{0.80}\cellcolor{Gray} & \textbf{6.30}\cellcolor{Gray} & 21.88 &  \textbf{0.84} \cellcolor{Gray}& 0.76\\
    \hline\hline
     rank scores [0-1]             & 0.000 \cellcolor{Gray}&  0.000 \cellcolor{Gray}& 0.004 & 0.000 \cellcolor{Gray}& 0.034\\
    \hline
    \end{tabular}

    \label{fig:testSetResultFiveSets} 
\end{table}
\vspace{-0.2cm}
\begin{table}
    \caption{Performance of top-5 methods among the 20 teams. The cells in gray shading indicate the best segmentation performance on each metric. The overall ranking is based on the average of the rank scores on each metric as shown in last row of Table \ref{fig:testSetResultFiveSets}. $\downarrow$ indicates that smaller value represents better performance.}
    \begin{tabular}{ |l | c | c | c | c | c | c |}
    \hline
    \textbf{Teams} &\textbf{Rank/score} &\textbf{DSC} & \textbf{H95$\downarrow$} & \textbf{AVD$\downarrow$} & \textbf{Recall} & ~\textbf{F1}~~\\
    \hline
     Ours &   ~\textbf{1}/0.038  & 0.80 \cellcolor{Gray}& 6.30 \cellcolor{Gray} & 21.88  & 0.84 \cellcolor{Gray} & 0.76\\
    \hline
     \emph{cian}       &  ~2/0.181  & 0.78 & 6.82  & 21.72  &  0.83  & 0.70\\
    \hline
     \emph{$nlp\_logix$}   & ~3/0.243  & 0.77 &  7.16 & 18.37\cellcolor{Gray} & 0.73  & 0.78\cellcolor{Gray}\\
    \hline
     \emph{$nih\_cidi\_2$} &  ~4/0.302 & 0.76 &  7.02 & 27.98 & 0.81 & 0.70\\
    \hline
     \emph{$nic-vicorob$} &  ~5/0.369  & 0.77 &  8.28 & 28.54 & 0.75  & 0.71\\

    \hline
    \end{tabular}

    \label{fig:testSetResultComparsion} 
\end{table}

\begin{table}
\scriptsize
\newcommand{\tabincell}[2]{\begin{tabular}{@{}#1@{}}#2\end{tabular}}
\renewcommand\arraystretch{1}
  \centering
  \caption{Brief description of top-five methods}\label{table:Table8}
\begin{tabular}{ll}
\toprule
  \textbf{Team Names}&\textbf{Brief Description of Methods}\\
\midrule
\emph{$sysu\_media(ours)$} & Fully convolutional network ensembles. \\
\midrule
\emph{cian}&Multi-dimensional gated recurrent units based on recurrent neural networks.\\
\midrule
\emph{nlp$\_$logix}&Two densely connected deep convolutional neural networks.\\
\midrule
\emph{$nih\_$cidi\_2}&Traditional deep fully convolutional neural network and graph refinement.\\
\midrule
\emph{$nic-vicorob$}&A cascade of three convolutional neural networks.\\

\bottomrule
   \end{tabular}
\end{table}


\subsection{Leave-one-subject-out evaluation on public training dataset} \label{leaveOfSubjectOut}
To test the generalization performance of our system across different subjects, we conducted an experiment on the public training datasets (60 subjects) in a leave-one-subject-out setting. Specifically, we used the subject IDs to split the public training dataset into training and validation sets. There were 60 different subjects available. In each split, we used slices from 59 subjects for training, and the slices from the remaining subject for testing. This procedure was repeated until all of the subjects are used as testing.

\begin{figure*}[t]
	\begin{center}
		\includegraphics[width=1.05\linewidth,height=0.65\linewidth]{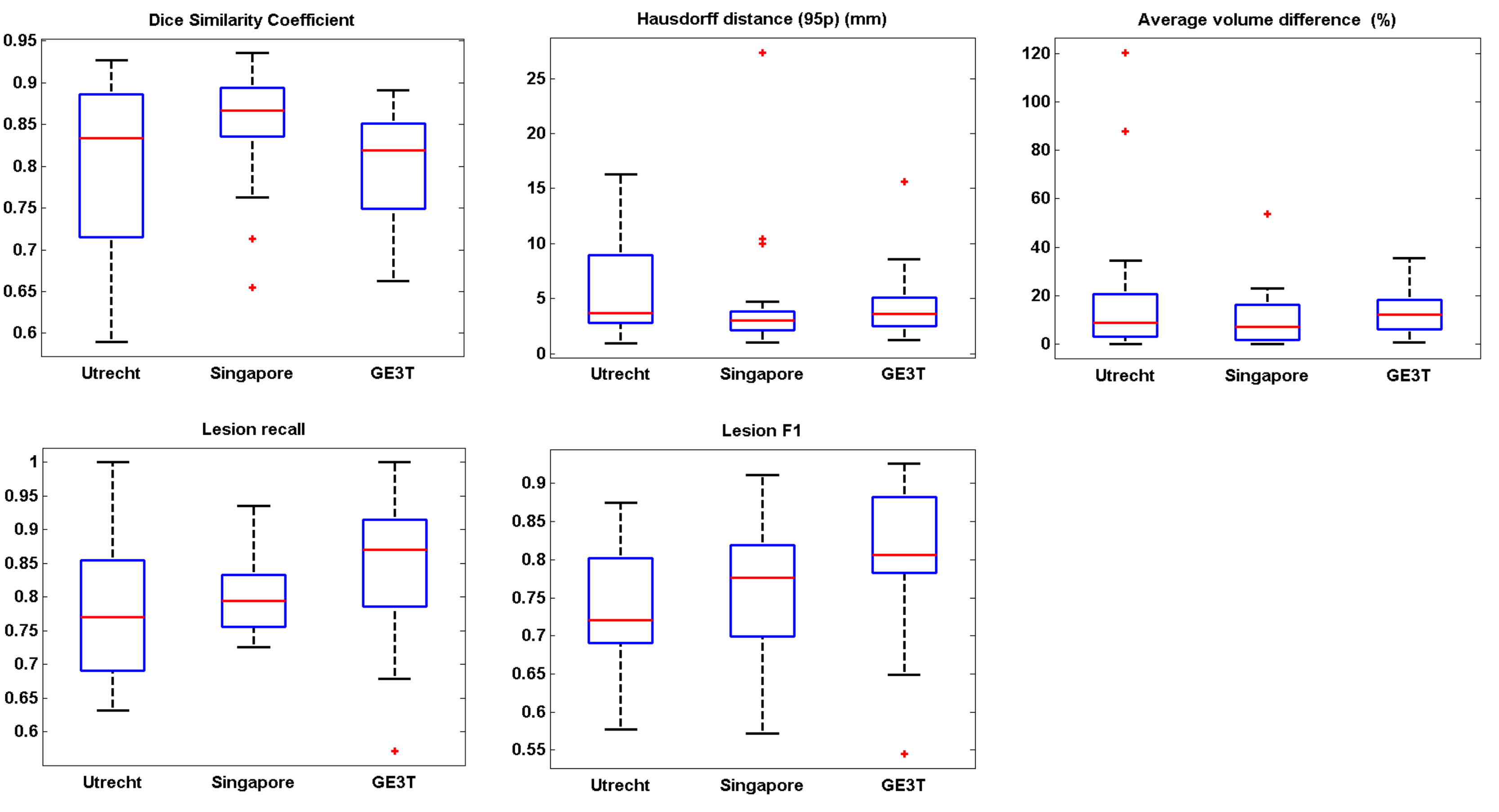}
	\end{center}
	\caption{Box plots of leave-one-subject-out evaluation on the public training data. Each box plot summarizes the segmentation performance on images from one scanner using one specific metric.}
	\label{fig:BoxLeaveOneOut} 
\end{figure*}

Figure \ref{fig:BoxLeaveOneOut} plotted the distributions of segmentation performances on scans from the three scanners, with each sub-figure showing performances using one of the five metrics. It could be observed that the segmentation performance on \emph{Utrecht} was relatively poor. A few outliers (hard examples) were found in \emph{Utrecht} which appeared to contain relatively more small lesions and blurred slices after checking the original slices and segmentation results. Section \ref{discussion} presents a further analysis of these outliers, revealing the challenge of WMH segmentation task.
In general, the averaged dice similarity coefficient, Hausdorff distance and lesion recall achieved by the proposed system on 60 cases were 87\%, 3.6mm and 85\%, respectively. This shows its effectiveness in aspects of overlapping, localization accuracy and overall lesion detection. Table S1 in the supplemental material reports extensive results allowing comparison on every case of the public training dataset.

\subsection{Cross-scanner Evaluation} \label{crossScanner}
To further evaluate the generalization performance to unseen scanners, firstly we presented a study of cross-scanner analysis on public training set containing 60 cases from three scanners.
Then we directly re-ranked and compared the cross-scanner segmentation performance of all teams' methods on the two unseen scanners.

For the cross-scanner analysis, we used the scanner IDs to split the 60 cases into training and test sets.
In each split, the slices of 40 subjects from two scanners were used as training set while the slices of 20 subjects from the remaining scanner were used for validation set. This procedure was repeated until all the scanners are used as validation set. For comparing the cross-scanner performance with other state-of-the-art methods, we calculated averaged performances of all teams on the two unseen scanners \emph{AMS GE1.5T} and \emph{AMS PETMR}. Then each team's ranking score was calculated using the same rank method introduced in Section \ref{evaluationAndRank}.

\begin{figure*}[t]
	\begin{center}
		\includegraphics[width=1.00\linewidth,height=0.60\linewidth]{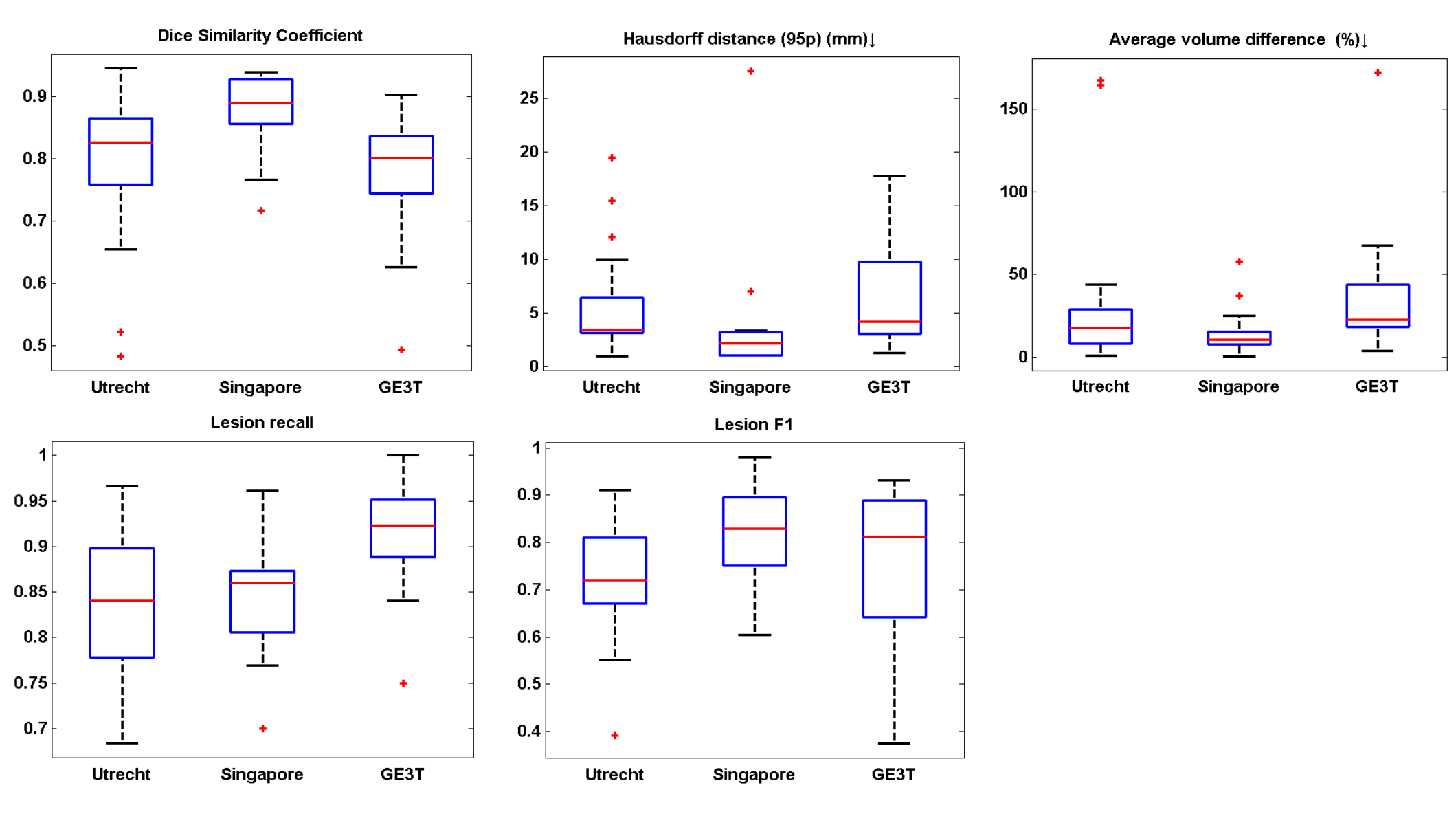}
	\end{center}
	\vspace{-0.2cm}
	\caption{Box plots of cross-scanner evaluation on the public training data. Each box plot summarizes the segmentation performance on subject from three testing scanners using one specific metric. For example, for box plot \emph{Utrecht} in the upper left figure, it shows the distribution of segmentation results on \emph{Utrecht} when training the model by using data from two other scanners - \emph{Singapore} and \emph{GE3T}.}
	\label{fig:BoxCrossScanner} \vspace{-0.2cm}
\end{figure*}

Figure \ref{fig:BoxCrossScanner} plots the distributions of segmentation performances on cases from each scanner being tested in turn, with each sub-figure showing performances using one of the five metrics.
In general, for every 20 cases from each of the three testing scanners in the cross-scanner evaluation, the segmentation result between each other was comparable, showing our system is robust to unseen scanners.
It could be observed that the segmentation performance on dataset \emph{GE3T} was relatively poor.
This could be explained that the voxel size of cases in \emph{GE3T} has a significant difference from that captured by two other scanners. Combination of modalities will be discussed in Section~\ref{influenceOfModalities}
Table \ref{fig:UnseenScannerResultComparsion} compares the segmentation performances of the top performing teams on two unseen scanners. Our method achieved, on average, the best Dice similarity coefficient and lesion recall of 74.5\% and 87\% respectively and runner-ups on other three metrics.

\vspace{-0.2cm}
\begin{table}
    \caption{Performance on two unseen scanners of top-5 methods among the 20 teams. The cells in gray shading indicate the best segmentation performance on each metric. The overall ranking is based on the average of the rank scores on each metric as shown in last row of Table \ref{fig:testSetResultFiveSets}. $\downarrow$ indicates that smaller value represents better performance.}
    \begin{tabular}{ |l | c | c | c | c | c | c |}
    \hline
    \textbf{Teams} &\textbf{Rank/score} &\textbf{DSC} & \textbf{H95$\downarrow$} & \textbf{AVD$\downarrow$} & \textbf{Recall} & ~\textbf{F1}~~\\
    \hline
     Ours &   ~\textbf{1/0.040}& 0.745 \cellcolor{Gray}&11.04  & 26.2& 0.87 \cellcolor{Gray}& 0.725\\
    \hline
     \emph{$nih\_cidi\_2$} &  ~2/0.234    & 0.705 & 9.745\cellcolor{Gray}& 21.94\cellcolor{Gray}&  0.79  & 0.685\\
    \hline
    \emph{cian} & ~3/0.264   & 0.745 \cellcolor{Gray} &14.10 & 28.425 & 0.82  & 0.665\\
    \hline
     \emph{$nic-vicorob$} &  ~4/0.374   & 0.715 &  13.53 & 56.31 & 0.815 & 0.62\\
    \hline
    \emph{$nlp\_logix$}&  ~5/0.408    & 0.685 &  12.98 & 27.9 & 0.665  & 0.73\cellcolor{Gray} \\

    \hline
    \end{tabular}
    \label{fig:UnseenScannerResultComparsion} 
\end{table}

\section{Discussion} \label{discussion}
In this section, we further present relevant results obtained on the training data and that impacted on our design choices.
\subsection{Why choose 2D architecture}
\begin{figure*}[t]
	\begin{center}
		\includegraphics[width=1\linewidth,height=0.48\linewidth]{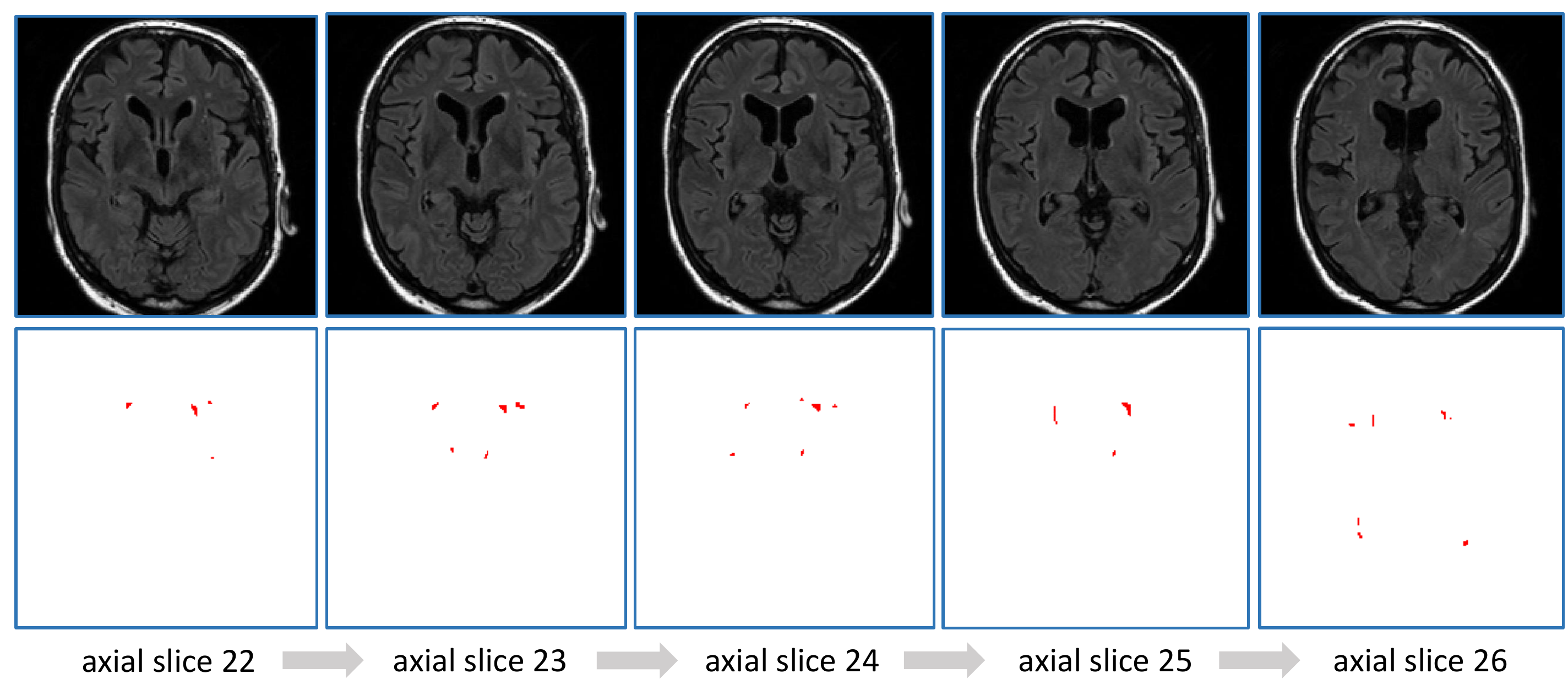}
	\end{center}
	\caption{Case 11 from the public training set shows the high discontinuity. From top to down, slices and corresponding ground-true segmentation maps. From left to right: axial slices from 22 to 26 and the corresponding ground truth.}
	\label{fig:incontinuity} 
\end{figure*}
It is noted that there exist several 3D convolutional network architectures for brain tumor segmentation \citep{kamnitsas2017efficient, havaei2017brain}. The main motivation of employing 3D architectures is to extract rich spatial and contextual information from tumor/lesion tissue volume. However, in case of WMH segmentation, small lesions with high discontinuity and low contrast are commonly found, which contain poor spatial and contextual information. Furthermore, the imaging resolution along \emph{z}-direction of the contest images is rather poor, and there exists large variation of spatial resolution as shown in Table \ref{table:Table2}, which further restricts the use of 3D deep learning models. Figure \ref{fig:incontinuity} shows the case 11 in dataset \emph{Utrecht}, in which small lesions with discontinuity characteristic are observed.
Therefore a 2D architecture is chosen for this challenge to explore the texture information at slice level, and to drastically reduce the computational complexity. Data augmentation further equips the 2D model with desired invariance and robustness. It should be acknowledged that, when large clinical datasets are available in future, 3D architectures might help to improve the segmentation performance.

\subsection{Analysis of U-Net hyper parameters} \label{parameters}
An appropriate parameter setting is crucial to successful training of deep fully convolutional networks. Here we mainly discuss some hyper parameters including the number of epochs, size of batch training and learning rate.

\begin{figure*}[t]
	\begin{center}
		\includegraphics[width=0.99 \linewidth,height=0.64\linewidth]{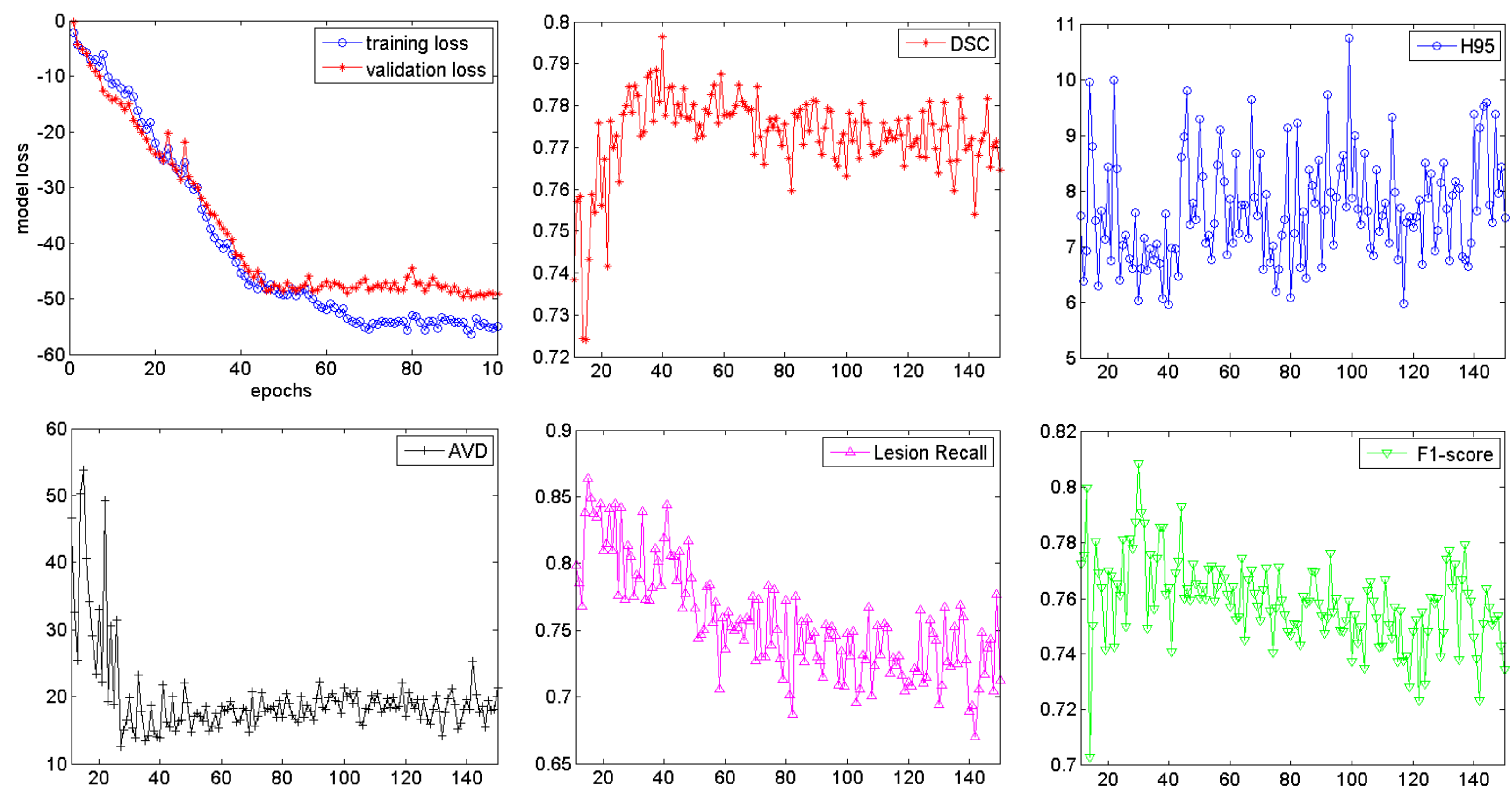}
	\end{center}
	\caption{Curves of training and validation loss and segmentation performance of each metric over epochs. 
          }
	\label{fig:loss} 
\end{figure*}

We selected the number of epochs for stopping training by contrasting training loss and validation loss over epochs. We split the public training dataset into a training set and a validation set by randomly picking 80\% and the remaining 20\% cases from each scanner respectively. Thus in total, the models were trained on 48 cases and validated on 12 cases. Figure \ref{fig:loss} shows the curves of training and validation loss over 100 epochs.
It could be observed that the validation loss did not show a descending trend at around 50 epochs. The reason to choose 50 epochs rather than a higher one is 1) to avoid over fitting on the training data, and 2) keep low computational cost.


The size of batch and learning rate have a large influence on the stability of the training process.
To our empirical observation, if the learning rate was set to values bigger than 10$^{-3}$, the training loss would be suddenly reaching to nearly 0 (i.e., the worst performance) at some beginning epoch and would remain not updating the training loss. Both of the batch size and learning rate directly influence the magnitude of the gradient and sometimes will lead to a gradient exposure issue.
Therefore the batch size was set to 30 and learning rate was set to 0.0002 throughout all of the experiments.


\subsection{Influence of imaging modalities} \label{influenceOfModalities}

\begin{figure*}[!ht]
\vspace{-2 cm}
	\begin{center}
		\includegraphics[width=1\linewidth,height=1.15\linewidth]{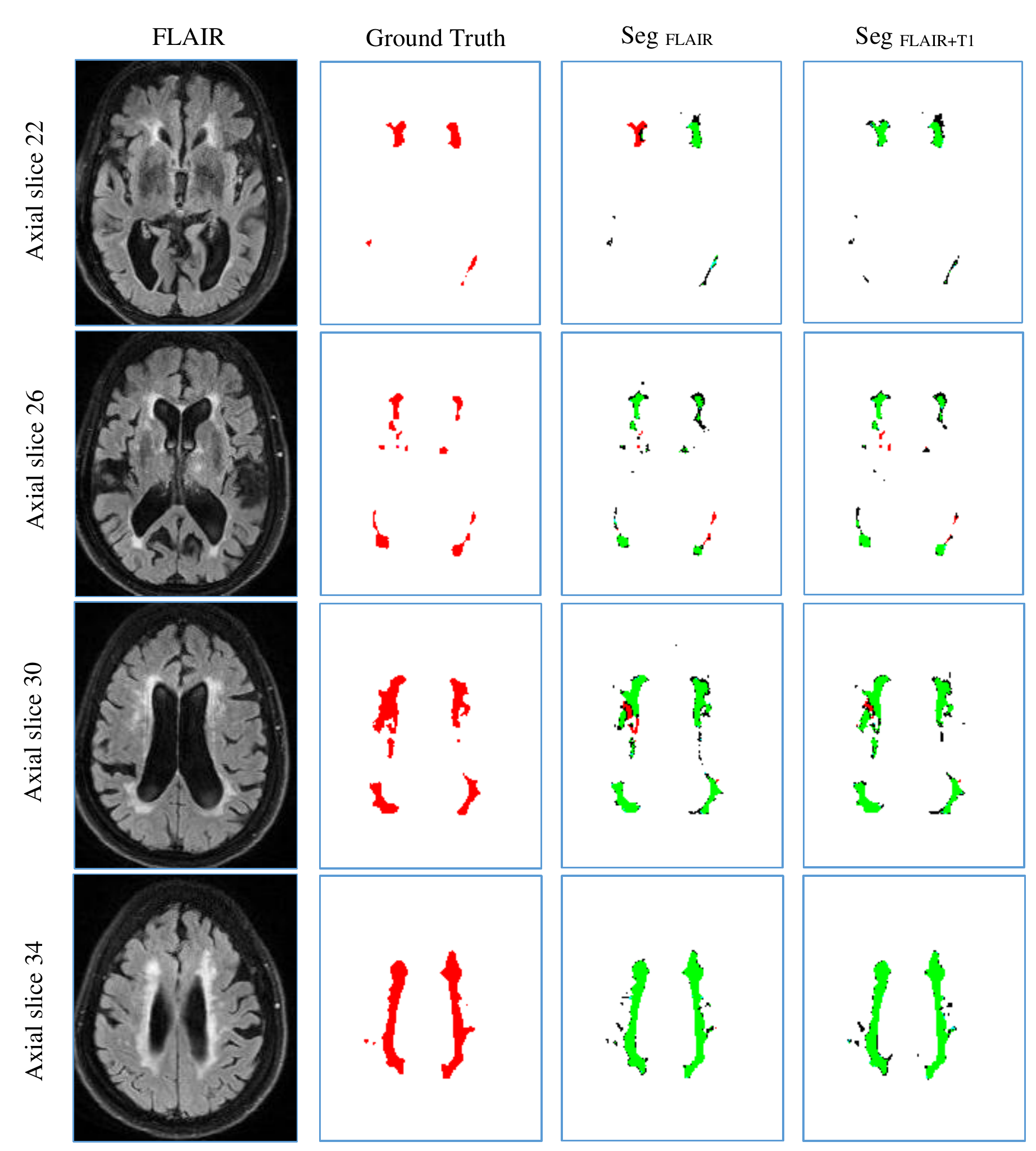}
	\end{center}
	\caption{Segmentation result on \emph{Singapore 34}. From top to bottom: four axial slices of the same subject. From left to right: FLAIR MR images, the associated ground truth, segmentation result using FLAIR modality only and segmentation result using FLAIR and T1 modalities. In column Seg$_{FLAIR}$ and Seg$_{FLAIR+T1}$, the green area is the overlap between the segmentation maps and the ground-truth, the red pixels are the false negatives and the black ones are the false positives. (Best viewed in colour)}
	\label{fig:comparionModality} 
\end{figure*}
The T1 modality is known to provide a good contrast between the healthy tissues of the brain while FLAIR sequences are widely used to distinguish pathologies present in the white matter. Based on this, we assumed that these two modalities can provide complementary information for segmenting WMH. According to previous work \citep{dyrby2008segmentation}, a combination of FLAIR and other modalities significantly improved the segmentation performance than using FLAIR alone. However, whether this combination improves the generalization capability to unseen scanner, has not been clearly investigated.
We therefore analysed and presented a novel study for comparison in a cross-scanner-evaluation manner.

Table S2 to Table S4 in supplemental material report extensive results. They show that the combination of FLAIR and T1 slightly outperformed FLAIR alone on most of the metrics, suggesting T1 modality could provide useful information for detecting WMH.
In Figure \ref{fig:comparionModality} we showed the segmentation results of a case from \emph{Singapore} tested by the model trained on \emph{Utrecht} and \emph{GE3T}. We observed that some false negatives were removed by using the combination of FLAIR and T1 after comparing the column \emph{Seg$_{FLIAR+T1}$} and \emph{Seg$_{FLIAR}$}, suggesting T1 provided complementary information on judging WMH. We further performed Wilcoxon signed rank test on the 60 cases. The improvements on H95 and F1-score were significant, giving p-values smaller than 1 $\times 10^{-4}$.

\subsection{Influence of data augmentation} \label{EffectsDataAugmentation}
The intention of data augmentation is generating training samples with different distributions to teach network learning desired invariance and robustness. We evaluated this technique using the cross-scanner evaluation as discussed in Section \ref{influenceOfModalities}. The same experimental setting was used.

\begin{figure*}[!ht]
	\begin{center}
		\includegraphics[width=1\linewidth,height=1.0\linewidth]{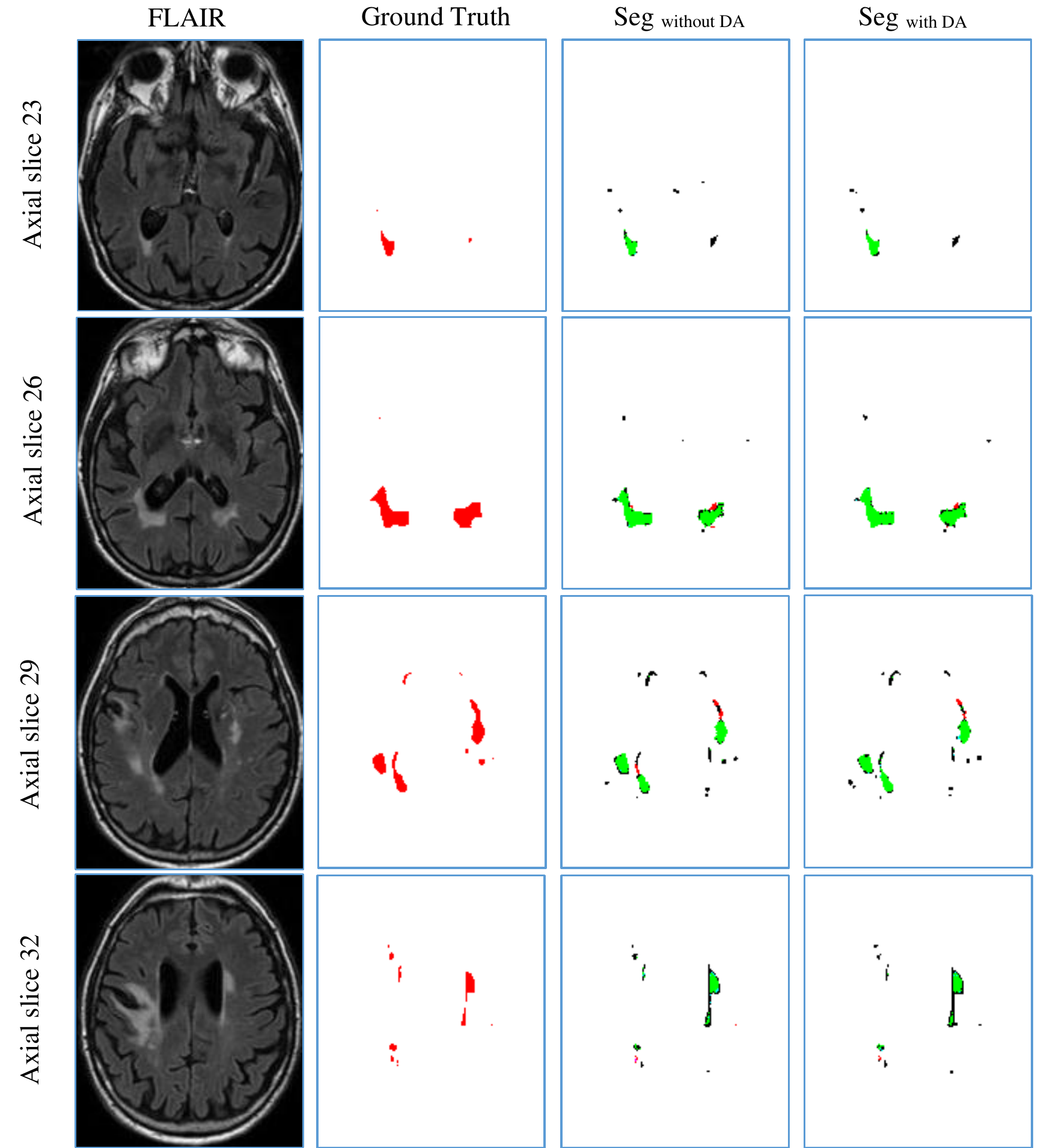}
	\end{center}
\caption{Sample segmentation result on \emph{Utrecht 04}. From top to bottom: four axial slices of the same subject. From left to right: FLAIR MR images, the associated ground truth, segmentation result without using data augmentation and segmentation result with data augmentation. In column Seg$_{without DA}$ and Seg$_{with DA}$, the green area is the overlap between the segmentation result and the ground truth, the red ones are the false negatives, and the black ones are the false positives. (Best viewed in colour)}
\label{fig:comparisonDataAugmentation}
\end{figure*}

Table S5 to Table S7 in supplemental material report extensive results. They show that using data augmentation slightly improved segmentation results on most of the metrics. Figure \ref{fig:comparisonDataAugmentation} shows the segmentation results of a case from \emph{Utrecht} tested by the model trained on \emph{Singapore} and \emph{GE3T}. We observed that some false positives with small volumes were removed by employing data augmentation after comparing the column \emph{Seg$_{without DA}$} to \emph{Seg$_{with DA}$}, suggesting the model achieved robustness to small lesions.
We further performed Wilcoxon signed rank test on the 60 cases. The improvements on H95, Recall and F1-score are statistically significant, giving p-values smaller than 1 $\times 10^{-4}$.

\subsection{Adaptability to different scanners}
To ensure the usability of the proposed system in real world practice, which involves imaging data from various scanners and protocols, we evaluated its adaptability to imaging data across scanners.
Extensive experiments were conducted by comparing the segmentation performances between models trained on either a single scanner or multiple ones.

Firstly, three sub-datasets from three scanners were evaluated independently. For example, 20 subjects from \emph{Utrecht} were split into training set and test set, and each subject was evaluated using the leave-one-subject-out evaluation introduced in Section \ref{leaveOfSubjectOut}.
Then the segmentation performance on each subject was compared to the one achieved by model trained on \textit{additional} data from other two scanners. This comparison allows us to see the adaptability of the system.

\begin{figure*}[!ht]
\vspace{-2.5cm}
	\begin{center}
\includegraphics[width=0.93\linewidth,height=1.25\linewidth]{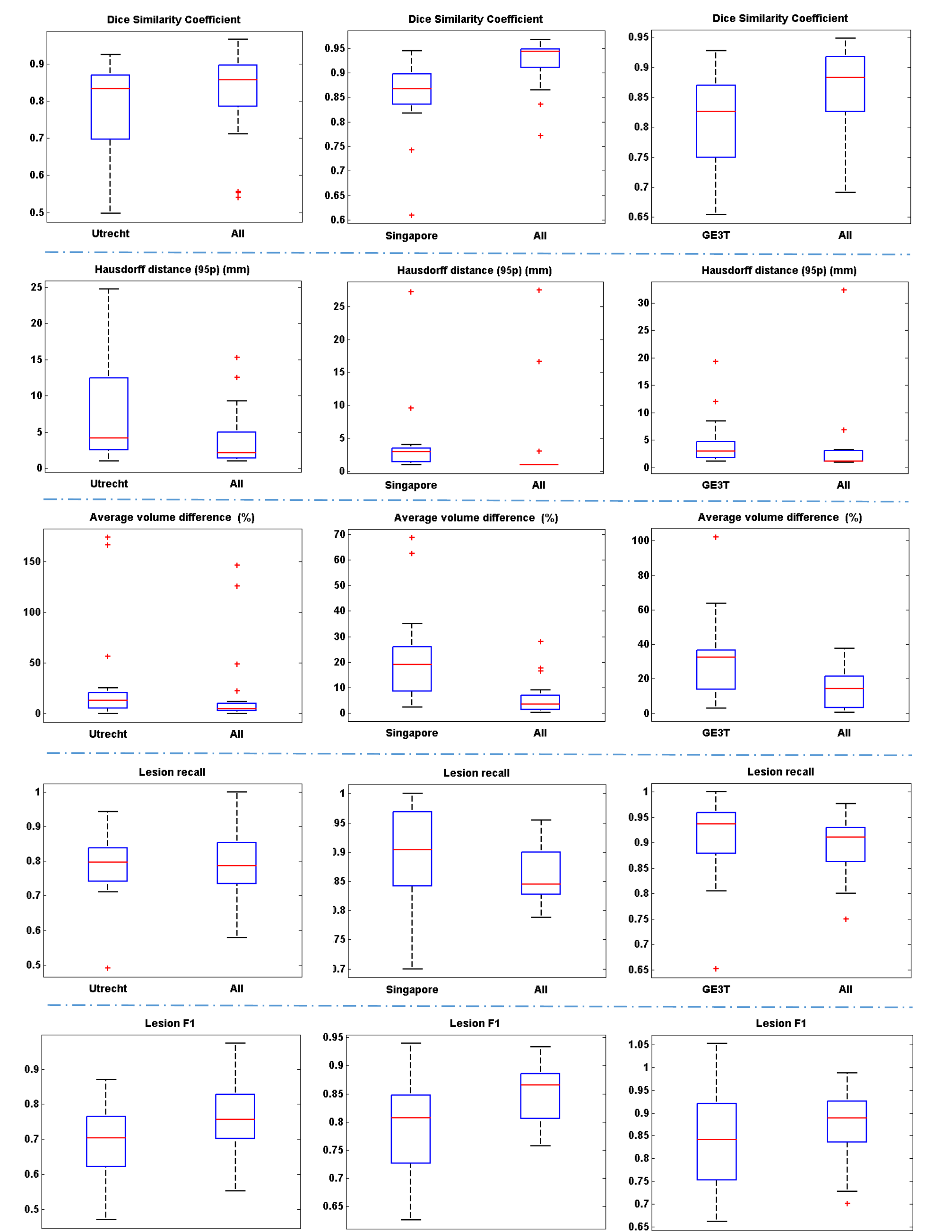}
	\end{center}
	\caption{Box plots of model adaptability evaluation. For example, the box plot in the left of first row shows two dice score distributions generated by two models trained on \emph{Utrecht} only and \emph{Utrecht} with additional data from other two scanners, respectively. From top to down: comparison of segmentation result on five metrics respectively. From left to right, comparison of segmentation result on \emph{Utrecht}, \emph{Singapore} and \emph{GE3T} respectively.}
	\label{fig:BoxAdaptablity} 
\end{figure*}

Figure \ref{fig:BoxAdaptablity} shows box plots of performances on each dataset.
Interestingly, we observed that, on four metrics - \emph{dice similarity coefficient, Hausdorff distance (95p), average volume difference and lesion F1-score}, the model trained on three scanners achieved significant improvement over the one trained on single scanner. However, on \emph{lesion recall}, the model trained on single scanner gained slightly better segmentation performance. This was due to the decrease of the number of undetected small lesions.
We concluded that the network trained on the larger data set that included cases obtained from different scanners shows better prediction performance, but at the cost of a sensitivity towards small lesions that were still detected best by networks trained on scanner- or sequence-specific data.

\subsection{Effect of the size of ensembles} \label{ensembles}
Ensemble learning aims at aggregating  different models to boost the segmentation performance.
The optimal size of an ensemble, i.e., how many models in the ensemble are needed, still remains an open issue and, as in many related ensemble learning task, a task specific parameter that needs to be optimized.
To this end, we evaluated how the segmentation performance behaves over the number of ensemble models.
We split the public dataset into training set and validation set by randomly picking 80\% and 20\% cases from
each scanner respectively. The models were trained on 48 cases and validated on 12 cases. Then the segmentation performance on 12 cases were averaged on each evaluation metric. For each model with different size of ensembles, the training process was repeated five times and the segmentation results on the validation set were averaged.

\begin{figure*}[!ht]
	\begin{center}
		\includegraphics[width=1.0\linewidth,height=0.48\linewidth]{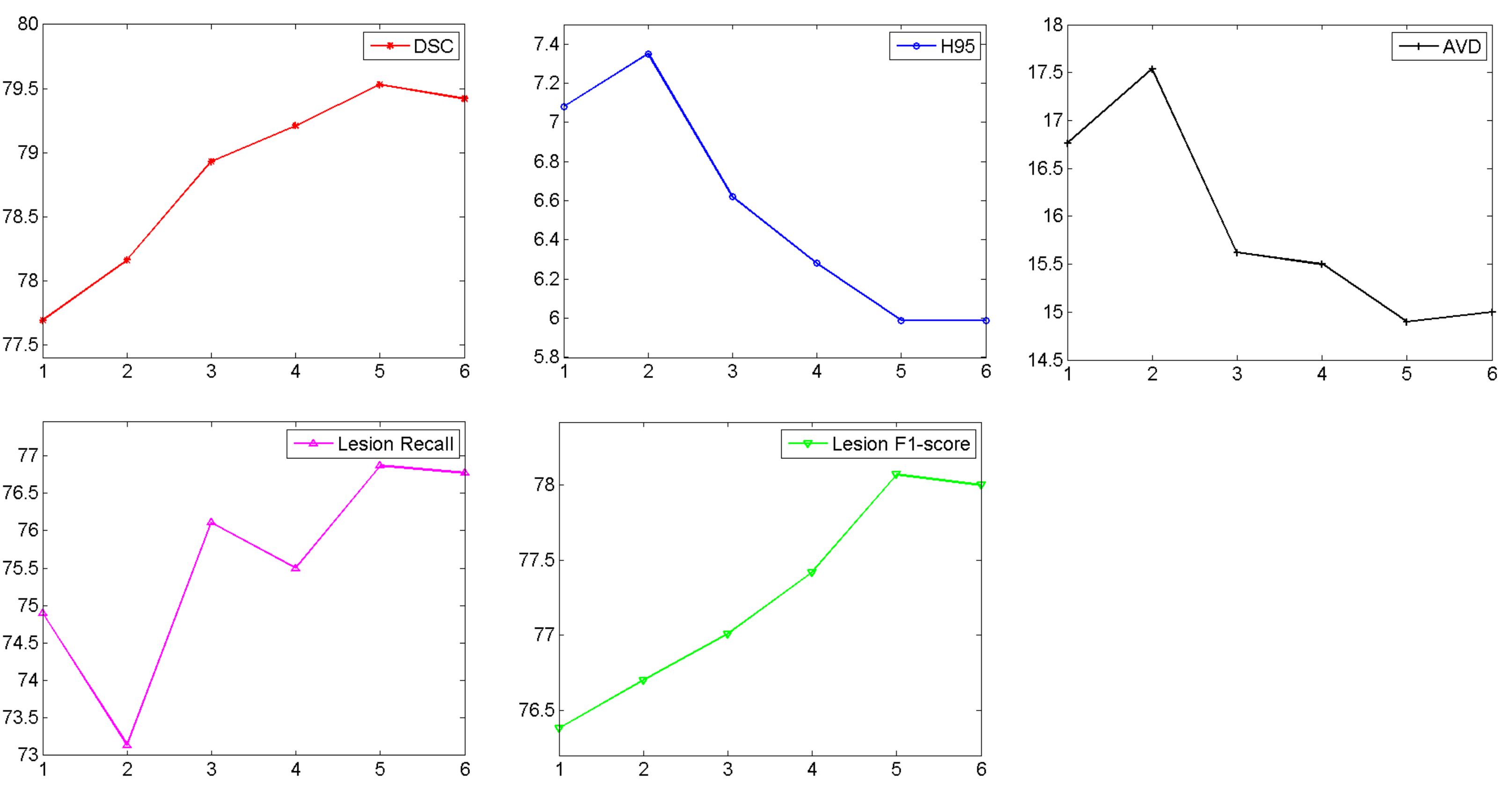}
	\end{center}
	\caption{Segmentation performance on validation set w.r.t ensemble size. The horizontal axis represents the number of models in the ensemble. We used an ensemble of three models in our final submission to the challenge.}
	\label{fig:numOfEnsembleModels} 
\end{figure*}

\begin{figure*}[!ht]
	\begin{center}
		\includegraphics[width=1.0\linewidth,height=0.48\linewidth]{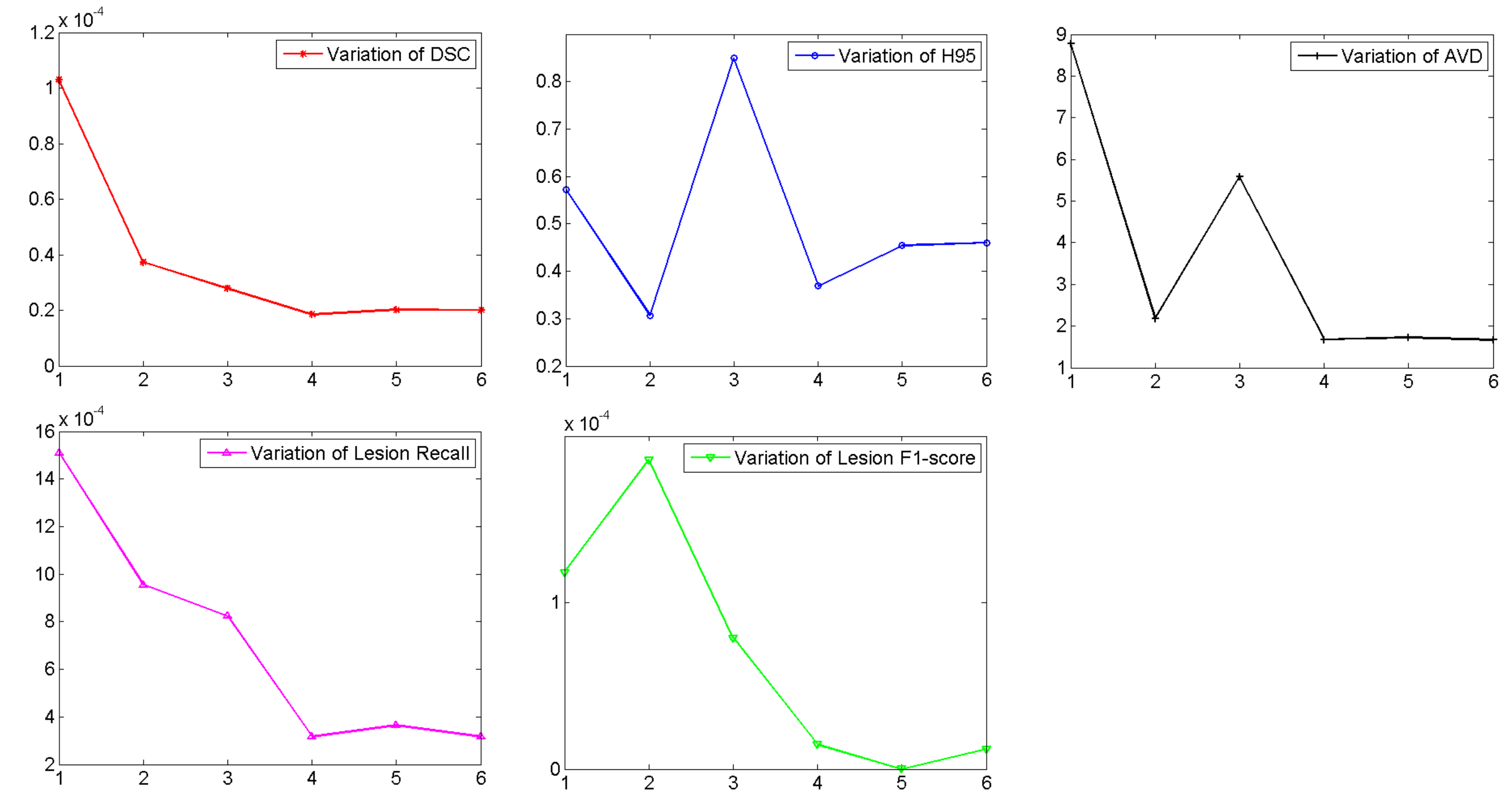}
	\end{center}
	\caption{The standard deviation of segmentation performance on validation set w.r.t ensemble size.
We observed that the variation of segmentation performance  was reduced when the size was increased.}
	\label{fig:EnsembleModelsVar} 
\end{figure*}

\begin{figure*}[!ht]
	\begin{center}
		\includegraphics[width=1\linewidth,height=0.65\linewidth]{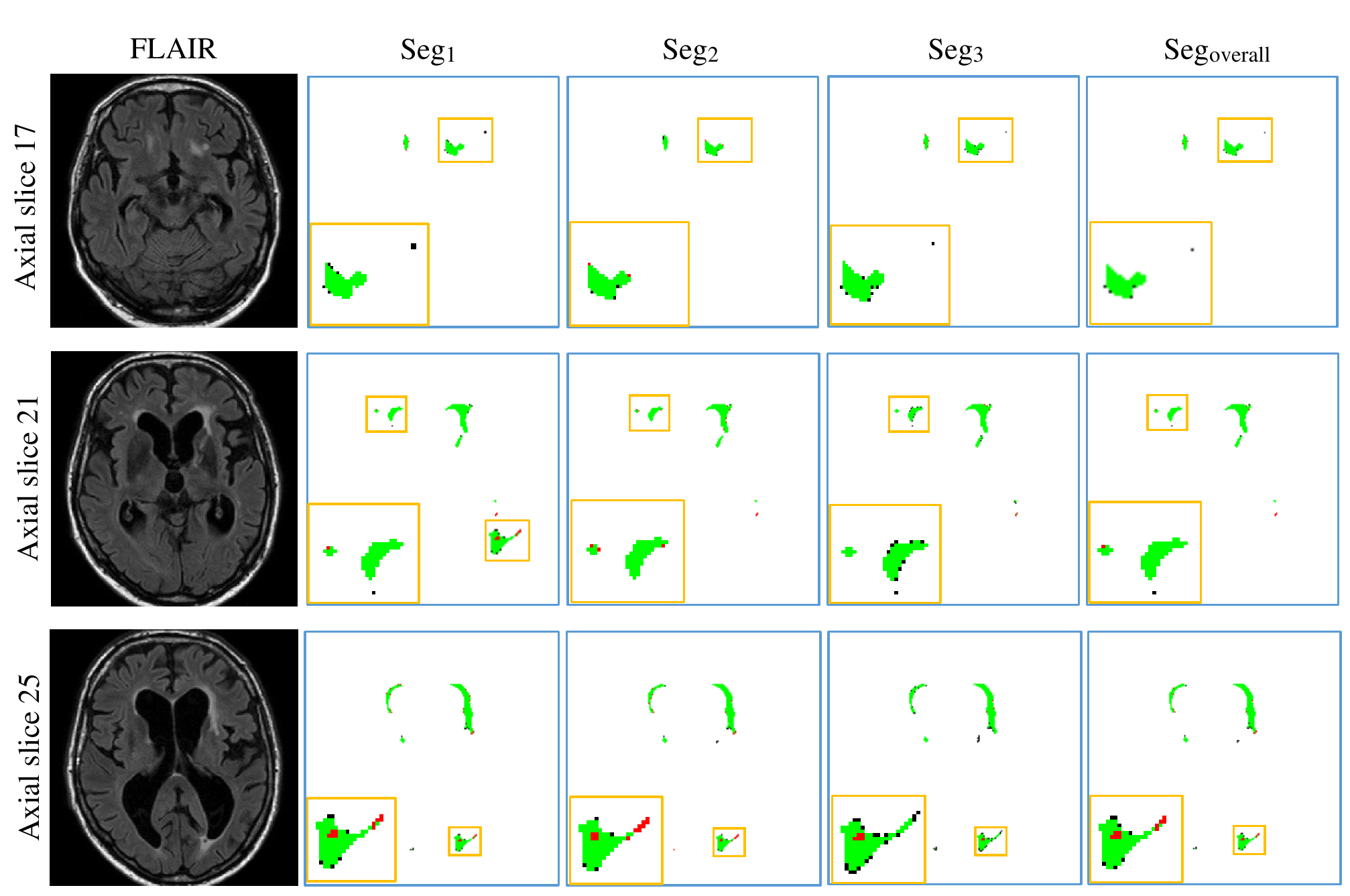}
	\end{center}
	\caption{Detailed segmentation results of three models and the ensemble. Columns \emph{Seg$_{1}$}, \emph{Seg$_{2}$}, \emph{Seg$_{3}$} and \emph{Seg$_{overall}$} represent the segmentation result generated by \emph{model 1}, \emph{model 2}, \emph{model 3} and their \emph{ensemble}. The green area in column Seg$_{1}$, Seg$_{2}$, Seg$_{3}$ and Seg$_{Overall}$ is the overlap between the segmentation result and ground truth. The red ones are the false negatives while the black ones are the false positives. For better visualization, the regions inside the smaller yellow bounding box are zoomed into the larger bounding box.}
	\label{fig:resultsEnsembles} 
\end{figure*}

Figure \ref{fig:numOfEnsembleModels} shows the curves of segmentation performance on five metrics w.r.t different ensemble size.
It could be seen that (1) the ensemble with three or more models clearly outperformed the ensemble of only one model on all of the five metrics. The improvement of ensemble model with size $5$ over one with size $3$ is statistically significant on five metrics, all with small p-values;
(2) when the size was further increased, performance tended to saturate and minor improvements in some of the measures came at the cost of small decreased in others.
Figure \ref{fig:EnsembleModelsVar} shows standard deviation of segmentation performance between five repeated trained models w.r.t different ensemble size. It could be observed that the variation of segmentation performance was reduced on the main evaluation metrics when the size of ensemble was increased. It demonstrated that the ensemble model can not only boost the segmentation performance but also guarantee a robust segmentation result.
Figure \ref{fig:resultsEnsembles} shows a case segmented by three individual models and their ensemble. 
We observed that three models trained with different weights initializations and shuffled data generated significantly different result on boundary and small lesions. And the model ensemble avoided the worst segmentation result.


\subsection{Statistical Analysis} \label{sec:analysis}

\subsubsection{Contribution of Each Component}

We investigated in depth the contribution of each component using statistical analysis.
Specifically, the performance of the proposed framework with and without a specific component was compared statistically as detailed below. For each of these comparisons, the public training dataset (from 60 patients) was first split into a training set and a validation set with a ratio of 4:1, resulting in a set of 48 training cases and a set of 12 validation cases.
Then the proposed framework without a specific component was trained on the 48 training cases and evaluated on each of the 12 validation cases w.r.t each of the five organizer-provided evaluation metrics.
The same protocol was also aplied to evaluate the complete proposed framework (i.e., without removing any component).
Then for each metric, Wilcoxon signed rank test was adopted to test the statistical significance of the difference between the proposed framework \textit{with} and \textit{without} a specific component based on their validation performance.
Since the comparisons were under a setting of multiple hypothesis testing, the p-values obtained for those five metrics were further adjusted by controlling the false discovery rate (PDR) for these hypothesis tests using the procedure proposed by \cite{benjamini1995control}.
Table \ref{Table:contribution} summarizes the contributions of each component in the framework as well as PDR-adjusted p-values of the test.
It could be observed that \textit{preprocessing}, \textit{data augmentation} and \textit{ensemble model} have consistent improvements on all of the five metrics. In particular, all the improvements of using data augmentation show statistical significance with very small p-values. On two metrics (H95 and AVD), the improvements of preprocessing are statistically significant. Similarity, the use of ensemble improves the performances on all the five metrics, among which, three (DSC, H95, AVD) are statistically significant. The use of the two modalities improves the performances on four metrics although no improvement was observed on AVD metric.

Overall,  the combination of these framework components helps build the state-of-the-art WMH segmentation system and differentiates our entry from other entries in the WMH segmentation competition.

\begin{table}
    \caption{The contribution of each component in the framework. \textit{p-val} denotes the adjusted p-value after controlling false discovery rate, and its bold face indicates statistical significance. \textit{IM} denotes the average improvement.} 
    \begin{tabular}{|p{2.1cm}|c|p{1.1cm}| c | p{1.1cm} | c | p{1.1cm} | c | p{1.1cm} | } 
    \hline
    \textbf{~}&\multicolumn{2}{|c|}{Preprocess} &\multicolumn{2}{|c|}{Data Aug.} &\multicolumn{2}{|c|}{Modalities}&\multicolumn{2}{|c|}{Ensemble}\\
    \hline
      ~ &IM & \textit{p-val} & IM & \textit{p-val}  & IM & \textit{p-val}  & IM & \textit{p-val}\\
    \hline
     \emph{DSC}&1.04\% & 0.1067& 1.38\%& \textbf{0.0030} & 0.62\% & 0.3393 & 1.98\%& \textbf{0.0115} \\
    \hline
     \emph{H95 (mm)$\downarrow$}&0.2&\textbf{0.0013}& 0.58& \textbf{0.0025} & 0.57 & \textbf{0.0013} & 0.95& \textbf{0.0025}\\
    \hline
     \emph{AVD$\downarrow$}& 2.15\% & \textbf{0.0013}& 3.02\%& \textbf{0.0025} & -0.96\% & 0.0013 & 2.29\%& \textbf{0.0025} \\
    \hline
     \emph{Recall} & 3.87\% & 0.1100 & 3.89\%& \textbf{0.0425} & 0.87\% & 0.4238 & 3.19\%& 0.9097 \\
    \hline
    \emph{F1-score} & 4.11\% & 0.1100 & 5.72\%& \textbf{0.0030} & 1.70\% & 0.3766 & 1.70\%& 0.5871 \\
    \hline

    \end{tabular}

    \label{Table:contribution} 

\end{table}

%
%
%

\vspace{-0.2cm}
\subsubsection{Best-Performing Model \emph{vs} Ensemble Model}
In practise, compared to the use of the ensemble for testing, one alternative approach is to selected a model from the ensemble, which performs the best on the validation set as the candidate model for testing. We refer this model as a \textit{best-performing} model. Here, we further compared the performances of best-performing model based on Dice loss and ensemble model.
Specifically, the public training dataset (60 cases) was split into a training set, a validation set and a test set with a ratio of 3:1:1, resulting in 36 training cases, 12 validation cases and 12 test cases.
We trained five models with different initializations, and selected the best-performing model based on the validation loss on the validation set. Then the performance of the best-performing model and the ensemble of the 5 models were compared on the \textit{test} set.
The averaged results on 12 test cases as well as the adjusted p-values of the Wilcoxon signed rank test after controlling the false discovery rate are shown in Table \ref{Table:bestPerforming}.
It shows that ensemble model outperforms single best-performing model on four metrics (significantly on Dice score and lesion F1-score).

\begin{table}
    \caption{Comparison of the best-performing model and ensemble model. The \edit{adjusted} p-values in bold indicate significant improvement achieved by ensemble model.}
    \begin{tabular}{ | c | c | c | c | c | c |}
    \hline
     \textbf{Models}&\textbf{~DSC~} & \textbf{~H95~$\downarrow$} & \textbf{~AVD~$\downarrow$} & \textbf{Recall} & ~~\textbf{F1}~\\
    \hline
     \emph{best-performing}   & 77.06\% & 7.87mm  & 16.78\%  & 71.60\%  & 72.99\%\\
    \hline
     \emph{ensemble model}& 78.80\% &  7.18mm & 18.92\% & 72.66\% & 77.29\%\\
    \hline\hline
     improvement  & 1.74\% & 0.71mm & -2.14\% & 0.84\% & 4.30\%\\
    \hline\hline
     p-value & \textbf{0.0015} &  0.20 & 0.0772 & 0.1496 & \textbf{0.0005}\\
    \hline
    \end{tabular}
    \label{Table:bestPerforming} 
\end{table}

\subsection{Computational Complexity}

All of the experiments were conducted on a GNU/Linux server running Ubuntu 16.04, with 32GB RAM memory.
The number of trainable parameters in the proposed model with two-channel inputs (FLAIR \& T1) is 8$,$748$,$609.
The algorithms were trained on a single NVIDIA Titan-Xp GPU with 12GB RAM memory. It takes around 180 minutes to train a single model for 50 epochs on a training set containing 10,000 images of size 200$\times$200 each. For testing, the segmentation of one scan with 48 slices by an ensemble of three models takes around 60 seconds using a Intel Xeon CPU (E3-1225v3) (without the use of GPU). In contrast, the segmentation per scan takes only 8 seconds when using a GPU.

\section{Conclusions}

In this paper we describe in detail our winning entry for MICCAI-2017 WMH Segmentation Challenge.
To investigate the contribution of  each component of our system,
we empirically study the effects of imaging modalities and data augmentation as well as ensemble size used in the system training that all contributed to the performance of our segmentation model. We found that (1) FLAIR and T1 imaging modalities provide complementary information to judge WMH;
(3) the proposed system shows good adaptability on various scanners and protocols;
(4) ensemble model helps to reduce over-fitting and boost segmentation results.
They are important factors to consider in building state-of-the-art WMH segmentation systems with good generalization capability.
The methods employed by the top-5 teams in the challenge are all deep-learning models, suggesting deep-learning techniques especially convolutional networks show high efficacy in WMH segmentation.
Although the segmentation results on 110 secret cases show its potential for real-world clinical use, the detection of small-volume WMH in MR images remains a challenging problem and is a future direction for the upcoming research in automated WMH segmentation.
Some interesting architecture which learns context information between slices \cite{chen2016combining} could be further investigated in future work. It will be interesting to discuss how segmentation difference between the algorithm and doctors will affect the clinical adoption, and how to address such a difference. This will need to test the algorithm in a clinical setting and get further feedback from radiologist and related therapist, which will be an interesting task in future work.
Note that our brain intensities are normalized based on all of the voxels within the brain in order to calibrate intensities across scanners. Since patients have varying amount of (hyper-intense) diseases, which may bias the mean intensities used in the normalization. To alleviate this bias, robust measures can be used, such as robust mean or median absolute deviance. Alternatively,  the lesion segmentation can be iterated and lesion areas identified in the first iteration are excluded in the normalization in the next iteration.
We make our \emph{Python} segmentation code available in \emph{GitHub}.

\section*{Acknowledgment}
We thank the MICCAI-2017 WMH Challenge organizer Dr.~Hugo~J.~Kuijf and the joint effort of the UMC Utrecht, VU Amsterdam, and NUHS Singapore for making the datasets and test results available for this research. We thank Haocheng Shen for the discussion during the challenge and presentation in the MICCAI BrainLes Workshop. The work was initialised when Hongwei Li was a visiting student in University of Dundee. The WMH contest submission was completed by a joint team from CVIP, Computing in University of Dundee and Sun Yat-Sen University. This work was supported in part by NSFC grant (No. 61628212), Royal Society International Exchanges grant (No. 170168), the Macao Science and Technology Development Fund under 112/2014/A3,  and Guangdong Program (No. 2014B010118003).

\section*{References}
\bibliographystyle{elsarticle-num-names}
\bibliography{egbib}

\end{document}